\documentclass[journal]{IEEEtran}
\usepackage{amsmath,amsfonts}
\usepackage{algorithmic}
\usepackage{algorithm}
\usepackage{array}
\usepackage[caption=false,font=footnotesize]{subfig}
\usepackage{textcomp}
\usepackage{stfloats}
\usepackage{url}
\usepackage{xurl}
\usepackage{verbatim}
\usepackage{graphicx}
\graphicspath{
  {images/setup/}
  {images/prediction/}
  {images/landing/}
}
\usepackage{tikz}
\usetikzlibrary{positioning,shapes.geometric,arrows.meta,calc}
\tikzset{
  startstop/.style={rectangle, rounded corners, draw=black, fill=gray!10, minimum width=2.0cm, minimum height=0.7cm, align=center},
  process/.style={rectangle, draw=black, fill=white, minimum width=3.0cm, minimum height=0.8cm, align=center},
  decision/.style={diamond, draw=black, fill=white, aspect=2, align=center, inner sep=1pt},
  arrow/.style={-Latex, thick}
}
\usepackage{cite}
\usepackage[hidelinks]{hyperref}
\hypersetup{
  pdftitle={Autonomous UAV Landing on a Heaving Platform Using Gated Recurrent Unit Prediction and Control-Barrier-Constrained Optimization},
  pdfauthor={Chiranjeev Prachand, Ritwik Shankar, Abhishek, Soumya Ranjan Sahoo, and K. S. Venkatesh},
  pdfsubject={Prediction-aided multirotor landing on a heaving platform},
  pdfkeywords={autonomous landing, control barrier functions, gated recurrent units, model predictive control, motion prediction, unmanned aerial vehicles, ship-deck landing}
}
\begin{document}

\title{Autonomous UAV Landing on a Heaving Platform Using Gated Recurrent Unit Prediction and Control-Barrier-Constrained Optimization}

\author{Chiranjeev~Prachand$^*$,
        Ritwik~Shankar$^*$,
        Abhishek,
        Soumya~Ranjan~Sahoo,
        and~K.~S.~Venkatesh
\thanks{$^*$These authors contributed equally to this work.}%
\thanks{R. Shankar and Abhishek are with the Department of Aerospace Engineering, Indian Institute of Technology Kanpur, Kanpur 208016, India.}
\thanks{C. Prachand, S. R. Sahoo, and K. S. Venkatesh are with the Department of Electrical Engineering, Indian Institute of Technology Kanpur, Kanpur 208016, India.}
\thanks{Supplementary video: \url{https://youtu.be/GB-uxwE1N9E}.}%
\thanks{This work has been submitted to the IEEE for possible publication. Copyright may be transferred without notice, after which this version may no longer be accessible.}}

\maketitle

\begin{abstract}
We present a prediction-aided architecture for landing a multirotor uncrewed aerial system on a heaving ship deck. A gated recurrent unit forecasts 6~s of heave using 40~s of history, predicted-peak selection identifies a heave peak for low-deck-velocity touchdown, and a control-barrier-function (CBF)-constrained model predictive controller generates the descent. Unsafe or infeasible approaches return to hover for reattempt. Experiments evaluated motion-capture (MoCap) input and onboard vision configurations. Mean forecast RMSE was 8.7~cm with MoCap and 8.8~cm with vision. All 20 trials, ten per configuration, achieved relative touchdown speed below 1~m/s. Eight MoCap and six vision trials landed on the first attempt; the remainder on the second. Mean absolute relative touchdown speeds were 0.626 and 0.656~m/s, and mean landing durations were 15.1 and 12.4~s, respectively. CBF rejection prevented inaccurate vision-predicted peaks from producing an unsafe continued descent. Comparable platform velocities at touchdown indicated favorable heave-phase selection. Results demonstrate feasibility of prediction-aided landing with abort-and-reattempt logic on an emulated heaving platform.
\end{abstract}


\section{Introduction}

\IEEEPARstart{A}{ship} deck undergoes stochastic motion whose severity is influenced by environmental conditions, including wind and waves, and by local bathymetry. A heaving ship deck can rise into a descending VTOL UAV, so a safe landing requires the vehicle to anticipate the deck's motion. However, both the onboard estimate of the deck's current height and any forecast of its future height carry errors that can lead to an unsafe touchdown. While precision landing of a multi-rotor Uncrewed Aerial Vehicle (UAV) is a widely researched topic in academia and industry, it is mostly limited to landing on a stationary platform with only roll and pitch motions \cite{saripalli_heli_landing, yang_stationary_landing} or a horizontally translating platform \cite{lima_vector_field_tethered_landing, niu_cbf_horizontal_landing, sankaranarayanan_cbf_horizontal_landing }. Landing a UAV on a heaving ship deck remains a challenging task.

Some works \cite{Nguyen2020HeavingLanding, lu2018shipboard} continuously track the ship deck height while reducing the error in vertical position. However, the UAV's limited vertical control bandwidth, motor response time, thrust constraints, and inertia introduce phase delay when tracking oscillatory deck motion, which can increase relative touchdown velocity. Unlike lateral tracking errors, vertical tracking errors directly affect impact conditions. Therefore, rather than following the instantaneous height of the deck, the descent should be timed using the deck motion history so that touchdown occurs near a favorable heave phase, preferably when the deck's vertical velocity is close to zero. For context, the nominal landing descent rate settings in widely used autopilots are, by default, 0.5 m/s in ArduCopter and 0.7 m/s in PX4. These commanded rates concern ordinary landing; for a heaving deck, the relevant touchdown measure is the UAV’s vertical velocity relative to the deck at first contact.

We address this timing problem with onboard visual estimation, a 6~s deck-heave forecast, and a controller that selects a predicted low-velocity heave peak for touchdown. During descent, a clearance constraint can reject an unsafe attempt and return the UAV to hover for another predicted peak. We evaluate the complete process on a heaving hardware platform, including touchdown velocity, rejected attempts, and landing time under vision and motion-capture sensing. The \href{https://youtu.be/GB-uxwE1N9E}{accompanying video} presents the system workflow and representative MoCap- and vision-input landing trials. This study evaluates the feasibility of the integrated architecture under the tested conditions.

Autonomous landing of vertical takeoff and landing (VTOL) UAVs has been approached using IR-LEDs, UWB, RTK-GPS, and vision. Vision is particularly attractive for landing on moving platforms because it provides relative deck pose using lightweight onboard sensing without requiring expensive localization infrastructure at the target~\cite{Xin_2022}. Early approaches relied on engineered landing patterns: Xu \textit{et al.}~\cite{Xu_2009} used a T-shaped shipboard target, while Zhang \textit{et al.}~\cite{Zhang_2019} demonstrated precise landing using fiducial-marker pose estimation. Subsequent developments in ARTag, AprilTag, and ArUco improved marker detection robustness and speed~\cite{Fiala,Wang_2016,aruco}, and marker-based landing was demonstrated on a moving ground vehicle in~\cite{Bhargavapuri_2019}. None of these studies addresses random heave motion of the landing surface.

Ship-deck landing is substantially more challenging due to coupled translational and rotational motion. Venugopalan \textit{et al.}~\cite{Venugopalan_2012} demonstrated landing on a kayak using colored targets, although the motion was primarily horizontal and ship-like roll, pitch, and heave were not quantified. Lin \textit{et al.}~\cite{Lin_2016} estimated four-DoF pose using an H-shaped marker and circle with partial-occlusion tolerance, but detection degraded below approximately 0.8~m and autonomous landing was not demonstrated. Wang and Bai~\cite{Wang_ship_deck_2017} combined LEDs with an H-shaped target on a three-DoF platform, but short-range LED visibility and limited roll, pitch, and heave restricted terminal landing evaluation.

Several later works improved pose estimation but retained similar limitations. Patruno \textit{et al.}~\cite{Patruno_2018} used an H-shaped pattern with concentric circles for six-DoF estimation but did not demonstrate autonomous landing. Polvara \textit{et al.}~\cite{Polvara_2018} demonstrated ARTag-based landing in Gazebo, although the simulated deck had small roll and no heave, and visual feedback was lost near touchdown. Sanchez-Lopez \textit{et al.}~\cite{Sanchez_Lopez_2013} used a neural network with a Stewart platform driven by stochastic ship-like motion, at the cost of training and computational requirements while still operating over a limited motion range. Similarly, a four-LED configuration in~\cite{langelaan_vision_2016} provided six-DoF pose estimation on a Stewart platform, but only one landing with low-amplitude roll, pitch, and horizontal motion was demonstrated, without heave.

More recent systems have increased platform mobility but still consider restricted deck dynamics. Lee \textit{et al.}~\cite{MOBLE_lee2023intelligent} mounted a Stewart platform on a ground vehicle and achieved accurate sway tracking using a rectangular visual target, although roll and heave remained limited. Shastry \textit{et al.}~\cite{Shastry} demonstrated ArUco-based tracking and landing on a swaying platform, but without roll or heave. Prachand \textit{et al.}~\cite{apna_paper} showed that fractal ArUco markers provide deck-relative pose over a wide altitude range without independently managing multiple markers, while Shankar \textit{et al.}~\cite{shankarVFS} subsequently demonstrated vision-based landing on a simulated deck undergoing combined roll, pitch, and sway.

Overall, existing vision-based ship-deck landing studies are commonly limited by restricted deck-motion amplitudes, incomplete motion models, loss of visual feedback during terminal descent, or lack of real-time hardware validation under coupled ship-like motion. Many studies focus primarily on six-DoF relative pose estimation or lateral tracking, while complete autonomous landing is either not demonstrated or evaluated only in simulation. Although simulation can incorporate sensor noise, hardware-dependent effects such as sensing and communication latency, actuator dynamics, structural inertia, and unmodeled disturbances are difficult to reproduce accurately unless explicitly modeled.

A further limitation is the restricted motion envelope of commonly used ship-deck emulators. Stewart platforms used in prior studies typically provide only a few centimeters of heave, commonly on the order of $\pm5$--$10$~cm, whereas the platform developed in this work provides up to 70~cm of heave travel. The larger workspace enables hardware experiments with substantially higher-amplitude motions representative of more realistic sea conditions. Alternative experiments using kayaks or small boats~\cite{Venugopalan_2012} provide physical validation, but artificially generated waves do not reproduce the magnitude, aperiodicity, and stochastic characteristics of open-sea ship motion.

The vision system introduces an additional challenge for terminal landing. Conventional fiducial markers generally require that all marker corners remain visible; therefore, a marker sized for long-range detection may leave the camera field of view during final descent, whereas a smaller marker may not be detectable from high altitude. Multiple-marker arrangements can extend the usable range, but require additional marker management and geometric transformations between marker and landing-platform reference frames. Therefore, this work adopts fractal ArUco markers~\cite{fractal}, in which nested markers of different scales provide continuous pose estimation over a large altitude range while maintaining a common reference geometry.

Prediction of ship motion and the occurrence of quiescent periods is itself an established research field with applications beyond UAV landing, particularly in naval engineering; a broad review is provided in~\cite{cademartori2023review}. Prediction methods can be broadly divided into model-based approaches, which incorporate vessel or wave dynamics, and time-series approaches that infer future motion directly from measurement history. Early work by Sidar and Doolin~\cite{Sidar1975CarrierMotion} showed that aircraft-carrier pitch and heave could be predicted in real time using Kalman-filter-based methods. Kusters \textit{et al.}~\cite{kusters2016futurewaves} later fused ship-motion measurements, a vessel model, and Doppler-radar observations of the ocean surface to predict ship motion several minutes ahead. Spectral approaches instead estimate dominant motion components using adaptive filters, observers, or Kalman filters. Abu Joub \textit{et al.}~\cite{abujoub2018unmanned} combined adaptive spectral prediction with LiDAR sensing for simulated UAV landing and reported fewer landing attempts, although at increased landing time. Monneau \textit{et al.}~\cite{monneau2020adaptive} used an adaptive notch filter to estimate frequency, amplitude, and phase, achieving a normalized root-mean-square error (RMSE) of 9.5\% with low computational cost. Triantafyllou, Bodson, and Athans~\cite{Triantafyllou1983ShipMotion} used Kalman filtering to estimate and predict heave, pitch, roll, sway, and yaw for VTOL landing applications, although no landing controller was developed, while Richter \textit{et al.}~\cite{Richter2014HeaveEstimation} proposed an adaptive IMU-based heave estimator that reduces phase error through online parameter adaptation but addresses only heave estimation.

Model-based prediction has also been integrated directly with landing control. Gupta \textit{et al.}~\cite{gupta2022landing} combined a Kalman-based ship-motion observer with model predictive control and demonstrated real-world landing on a platform mounted on a raft. However, the vertical motion was generated artificially and remained substantially more sinusoidal than ship motion in open-sea conditions. Hu \textit{et al.}~\cite{Hu2018TimeOptimalLanding} combined motion estimation, time-optimal trajectory generation, and tracking control for a vertically moving platform, using a UKF with a known multi-sinusoidal heave model for prediction. Such approaches provide efficient real-time prediction when an appropriate motion model is available, but their performance depends on how well the assumed periodic or parametric model represents the quasi-periodic and stochastic nature of actual ship motion.

Time-series methods can be further divided into statistical and machine-learning approaches. Among statistical predictors, autoregressive (AR) models and minor component analysis (MCA) are widely used. MCA-based prediction has been incorporated into UAV landing in~\cite{ngo2022variable,horn2015autonomous} and can achieve greater prediction accuracy than simpler spectral techniques~\cite{monneau2020adaptive}, although with higher computational cost; these studies were limited to simulation. AR models are computationally lighter and more effective over short prediction horizons. Patel \textit{et al.}~\cite{patel2021gpc} used AR prediction to identify quiescent periods and initiate landing when a sufficiently safe window was detected. AR-based predictions have also been used for shipboard path planning~\cite{pravitra2022shipboard}, while~\cite{scaled_experiments} coupled deck-heave forecasting with a quadratic-programming (QP) formulation for landing-trajectory generation and explicitly treated prediction as part of autonomous landing timing.

Machine-learning-based ship-motion prediction has gained increasing attention as computational hardware has improved. Wang \textit{et al.}~\cite{Wang2021-sn} predicted heave using a regularized back-propagation neural network, while recurrent architectures such as RNNs, LSTMs, and GRUs have become common for nonlinear time-series prediction. Keijzer~\cite{Keijzer2022ShipHeave} showed that LSTM networks can predict ship heave from motion history alone, without direct wave-excitation measurements, with stronger performance over shorter horizons. Del Aguila Ferrandis \textit{et al.}~\cite{Del_Aguila_Ferrandis2021-dd} compared LSTM, GRU, and conventional RNN models and reported superior prediction accuracy for LSTM. Lee \textit{et al.}~\cite{lee2023prediction} combined LSTM-based motion-history prediction with a CNN operating on spatio-temporal ocean-wave imagery obtained from marine radar. Geng \textit{et al.}~\cite{Geng2023EMDPSOLSTM} combined empirical mode decomposition with an adaptive PSO--LSTM predictor in a sliding-window framework to address nonlinear and nonstationary motion. Skulstad \textit{et al.}~\cite{skulstad2020hybrid} used a hybrid LSTM and vessel-dynamics model and reported improved 30-s-ahead prediction relative to a standalone LSTM.

Yang \textit{et al.}~\cite{Yang2025ReservoirComputing} used reservoir computing (RC) for real-time prediction of ship-motion and reported better performance than their tested GRU baselines. However, the comparison used only one relatively small, validation-selected GRU architecture per sea state, with a single GRU layer of 64 or 128 hidden units. Moreover, computation times were evaluated on a laboratory computer equipped with an Intel Xeon E5-2678 CPU and NVIDIA RTX 2080 Ti GPU, rather than UAV-compatible embedded hardware.

For autonomous landing, prediction accuracy alone is insufficient; the predictor must also operate at a rate compatible with closed-loop control. Spectral and low-order model-based methods generally have low computational cost and can support prediction rates of approximately 20--50~Hz, but their accuracy degrades when the actual motion of the deck deviates from the assumed periodic model. AR methods perform well over short horizons, typically a few seconds, but prediction quality deteriorates as the horizon increases. MCA and ML-based approaches can provide higher accuracy, but their computational requirements increase with input history, prediction horizon, and model size, creating a tradeoff between accuracy and real-time execution. Large neural networks may also require GPU-class hardware, which imposes significant mass and power penalties on small UAVs.

A further limitation of several landing approaches is their dependence on detecting and waiting for a quiescent period~\cite{abujoub2018unmanned,patel2021gpc}. Although this can improve touchdown safety, it may increase landing time and, therefore, reduce the endurance available for the UAV's primary mission in order to reserve adequate energy for landing. A more adaptable landing strategy would use short-horizon predictions to shape the descent trajectory and select a favorable touchdown phase without requiring a completely quiescent period to occur. This motivates an integrated architecture that combines robust vision-based relative-state estimation, real-time deck-heave prediction, trajectory generation, and safety-aware touchdown logic. In contrast to the model-based formulation of~\cite{Hu2018TimeOptimalLanding}, the present approach sacrifices guaranteed time optimality in favor of data-driven heave prediction, favorable-phase touchdown, and abort-and-reattempt capability under prediction uncertainty.

In piloted helicopter ship-deck landings, pilots do not simply follow the instantaneous deck motion but anticipate its future motion and time the final descent accordingly \cite{Vorwald2016-nb}. Prediction can likewise provide an autonomous controller with a future deck-motion estimate from which to select a low-velocity touchdown window and shape the descent. The present experiments assess the feasibility of this prediction-aided strategy rather than its performance advantage over a reactive alternative.

This paper proposes a heave-landing algorithm that uses augmented vision estimates from fractal ArUco markers and a Gated Recurrent Unit (GRU)-based recurrent neural network~\cite{GRU_original} to predict ship-deck heave. Based on these predictions, a landing window is estimated, and a descent trajectory is generated to land the UAV within it. To address prediction errors, a failsafe based on a control barrier function (CBF)~\cite{ames2017cbf} is implemented within the trajectory-generation module. The failsafe aborts an attempt when it detects insufficient clearance between the UAV and the platform, preventing the UAV from crashing into the ship deck platform. The main contributions of the paper are as follows:
\begin{enumerate}
    \item An integrated onboard perception--prediction--control architecture combining 6-s GRU heave forecasts, predicted-peak selection, and deck-relative MPC descent generation.
    \item A stopping-distance CBF condition and supervisory state machine that reject an infeasible or unsafe descent, return the UAS to hover, and permit a subsequent attempt.
    \item A hardware feasibility evaluation under controlled heave using MoCap- and vision-based configurations, with forecast, selected-window, attempt-count, physical first-contact, and landing-duration measurements.
\end{enumerate}
The remainder of the paper is organized as follows: Section~\ref{system_setup} provides an overview of the system setup. The prediction module is explained in Section~\ref{prediction_module}. Section~\ref{landing_algorithm} develops the failsafe landing algorithm, including the estimation of the landing window. The experimental validation and results are presented in Section~\ref{experiments}. Finally, the conclusions are discussed in Section~\ref{conclusions}.

\section{System Overview}
\label{system_setup}

\begin{figure}[t]
    \centering
    \includegraphics[width=\linewidth]{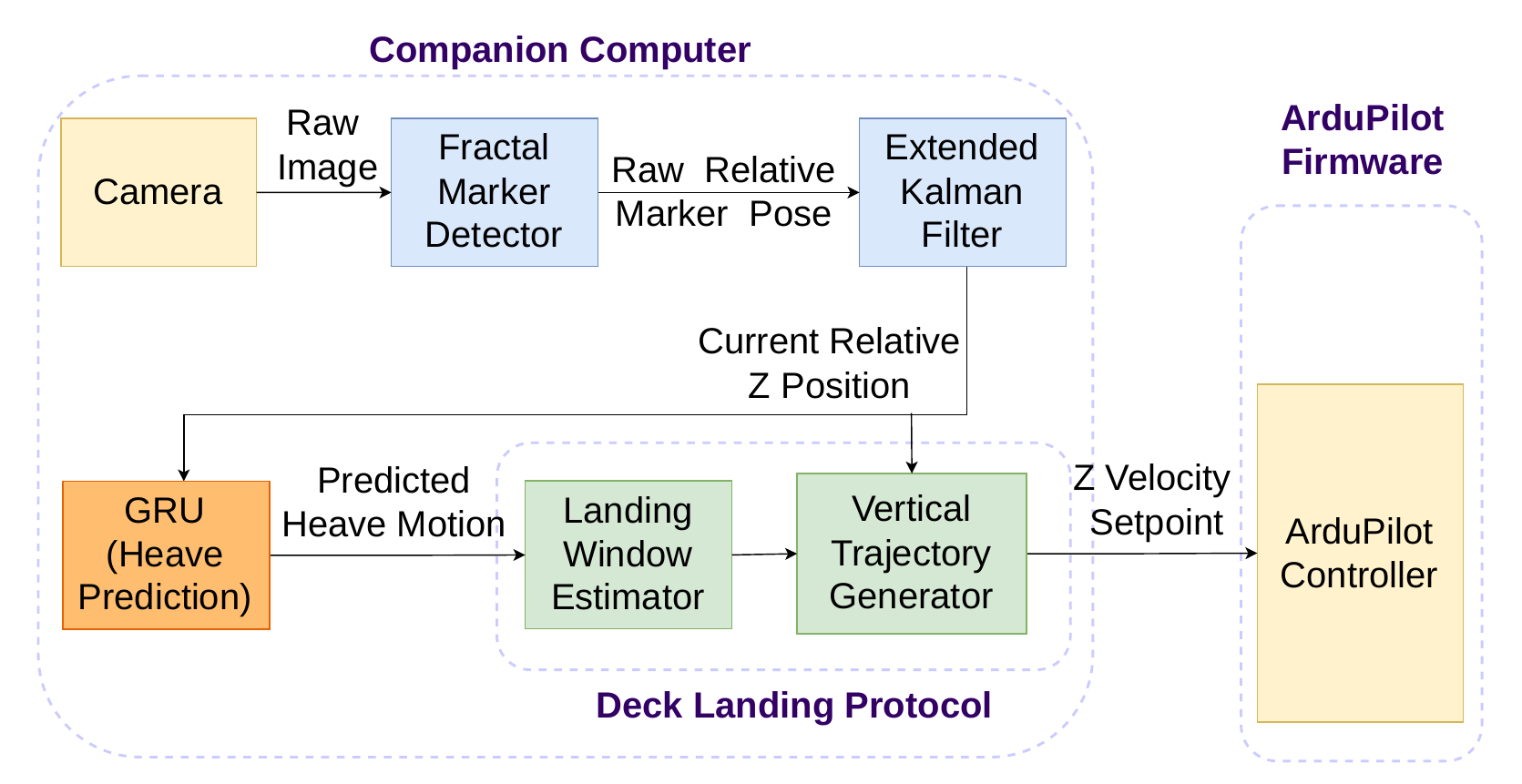}
    \caption{Perception--prediction--control architecture for heave-synchronized UAV landing.}
    \label{fig:heave_pipeline}
\end{figure}

The proposed system comprises an aerial vehicle, an actuated landing platform, and the perception--prediction--control pipeline in Fig.~\ref{fig:heave_pipeline}. A downward-facing camera detects a fractal ArUco marker~\cite{fractal} fixed to the platform and estimates the deck-relative pose. The calibrated camera-to-UAV transformation expresses this measurement in the UAV frame, after which a constant-acceleration extended Kalman filter rejects outliers, filters the relative position, and estimates relative velocity. A guarded affine correction compensates for the measured temporal lag in the vertical position and velocity estimates. The resulting relative vertical state is combined with the UAV state to reconstruct the heave motion of the platform. These perception and filtering stages provide the state input to the Gated Recurrent Unit (GRU), which predicts the future heave sequence. The landing-window estimator selects a low-velocity peak, and the vertical trajectory generator solves the constrained descent problem and sends the velocity setpoints to the flight controller. If the selected window becomes infeasible or violates the safety condition, the state machine aborts the descent, returns the UAV to hover, and evaluates a subsequent prediction. The hardware and test-specific parameters are reported in Section~\ref{subsec:experimental_setup}.

\subsection{Reference Frames and Notation}
\label{subsec:frames_notation}
Four right-handed frames are used. The inertial frame $\mathcal{F}_W$ has an upward-positive vertical axis $z_W$. The UAV frame $\mathcal{F}_U$ is attached to the vehicle, the camera frame $\mathcal{F}_C$ is attached to the downward-facing camera, and the platform frame $\mathcal{F}_P$ is attached to the center of the fractal marker at the landing surface. The camera measurements are transformed from $\mathcal{F}_C$ through $\mathcal{F}_U$ and expressed along $z_W$. Although the flight controller may use its native navigation convention internally, all vertical quantities in this paper use the upward-positive convention.

Let $z_U^W$ and $z_P^W$ denote the vertical positions of the UAV and platform expressed in $\mathcal{F}_W$. The measured vertical separation is
\begin{equation}
z_{\mathrm{rel}}=z_U^W-z_P^W,
\label{eq:relative_height_definition}
\end{equation}
which is positive when the UAV is above the platform. The vision-based platform-height estimate is therefore
\begin{equation}
\hat z_P^W=z_U^W-z_{\mathrm{rel}}.
\label{eq:platform_height_reconstruction}
\end{equation}
The controller later denotes the same geometric clearance between the UAV and the platform by $d$; thus, $d=z_U^W-z_P^W$, while $z_{\mathrm{rel}}$ denotes its measured counterpart. A superscript identifies the frame in which a quantity is expressed, a hat denotes an estimate or prediction, and the superscript $+$ denotes a future-horizon sequence. Table~\ref{tab:notation} summarizes the notation used throughout the prediction and landing formulations.

\begin{table}[t]
\centering
\caption{Reference frames and principal notation.}
\label{tab:notation}
\setlength{\tabcolsep}{3pt}
\begin{tabular}{@{}p{0.22\columnwidth}p{0.70\columnwidth}@{}}
\hline
\textbf{Symbol} & \textbf{Definition} \\
\hline
$\mathcal{F}_W$ & Inertial/world frame; $z_W$ is positive upward. \\
$\mathcal{F}_U$ & UAV-fixed frame. \\
$\mathcal{F}_C$ & Downward-facing camera frame. \\
$\mathcal{F}_P$ & Platform/marker frame at the landing surface. \\
$z_U^W,z_P^W$ & UAV and platform vertical positions in $\mathcal{F}_W$. \\
$z_{\mathrm{rel}}$ & Measured UAV--platform vertical separation. \\
$\hat z_P^W$ & Estimated platform height reconstructed from vision. \\
$d$ & Control clearance, $z_U^W-z_P^W$. \\
$v_U,v_P$ & UAV and platform vertical velocities. \\
$a_P,u$ & Platform acceleration and commanded UAV acceleration. \\
$\hat z_{P,k}^{W+}(i)$ & Platform-height prediction at horizon index $i$, issued at $t_k$. \\
$f_s$ & Prediction sampling frequency. \\
$N_h,N_p$ & GRU history and prediction-horizon lengths. \\
$N$ & MPC horizon length. \\
$N_w,N_s$ & Smoothing- and peak-search-window lengths. \\
$z_{\mathrm{peak}},T_{\mathrm{peak}}$ & Selected peak height and remaining time to the peak. \\
$v_r$ & Relative vertical velocity, $v_U-v_P$. \\
$d_{\min}$ & UAV-center clearance at landing-gear contact. \\
$a_{\mathrm{rel,max}},\gamma$ & Relative braking bound and CBF gain. \\
$\Delta t$ & Sampling interval. \\
\hline
\end{tabular}
\end{table}

\subsection{Vision-State Filtering and Lag Compensation}
\label{subsec:lag_compensation}
The transformed vision measurement is expressed using the upward-positive convention of Section~\ref{subsec:frames_notation} and SI units. A constant-acceleration multiplicative extended Kalman filter (MEKF) estimates the relative translational state, along with the marker orientation and angular velocity. Its vertical components are denoted by
\begin{equation}
\hat{\mathbf{x}}_{k}^{\mathrm{MEKF}}=
\begin{bmatrix}
\hat z_{\mathrm{rel},k}^{\mathrm{MEKF}} &
\hat v_{\mathrm{rel},k}^{\mathrm{MEKF}} &
\hat a_{\mathrm{rel},k}^{\mathrm{MEKF}}
\end{bmatrix}^{T}.
\label{eq:mekf_vertical_state}
\end{equation}
The MEKF uses the standard constant-acceleration prediction followed by a measurement correction. Position innovations exceeding $0.35$~m, or vertical innovations exceeding $0.12$~m, are rejected.

Filtering reduced measurement noise but left a systematic lag in the estimated vertical position and velocity. A separate constant-acceleration Kalman filter was therefore applied to the transformed vertical measurement. Its state
$\mathbf{x}_{k}^{L}=[z_k^{L},v_{z,k}^{L},a_{z,k}^{L}]^{T}$ is propagated using
\begin{equation}
\mathbf{F}(\Delta t)=
\begin{bmatrix}
1 & \Delta t & \Delta t^2/2\\
0 & 1 & \Delta t\\
0 & 0 & 1
\end{bmatrix}.
\label{eq:lag_filter_model}
\end{equation}
The process-noise covariance is $\mathbf{Q}_{L}=0.03\mathbf{I}_{3}$, and the scalar measurement-noise variance is $R_L=0.0005$. Vertical innovations greater than $0.12$~m are rejected. The filter is reinitialized from the current measurement after three consecutive rejections. An acceleration feature is obtained from the derivative of the filtered velocity and exponentially smoothed as
\begin{align}
\tilde a_k&=\tilde a_{k-1}+\alpha_k
\left(\frac{v_{z,k}^{L}-v_{z,k-1}^{L}}{\Delta t}
-\tilde a_{k-1}\right),\label{eq:lag_acceleration_feature}\\
\alpha_k&=1-\exp\left(-\frac{\Delta t}{0.10}\right).
\end{align}
The compensated relative position and velocity candidates are then calculated from
\begin{align}
z_{\mathrm{corr},k}
&=0.0062265+0.9769471z_k^{L}
+0.0338166v_{z,k}^{L}\nonumber\\
&\quad-0.0055090\tilde a_k,\label{eq:lag_corrected_position}\\
v_{\mathrm{corr},k}
&=0.0388227-0.0364110z_k^{L}
+0.7425399v_{z,k}^{L}\nonumber\\
&\quad+0.2268494\tilde a_k.\label{eq:lag_corrected_velocity}
\end{align}
The coefficients were fitted in a separate calibration experiment in which the marker remained stationary while the UAV moved vertically above it. Synchronized vision estimates were used as inputs, and measurements from a Qualisys motion-capture (MoCap) system provided the relative-state reference. The prediction-validation and landing datasets were not used to identify these coefficients.

The corrected candidates are accepted only when both estimates are finite, the image age is no greater than $0.85$~s, and
\begin{align}
\left|z_{\mathrm{corr},k}-\hat z_{\mathrm{rel},k}^{\mathrm{MEKF}}\right|
&\leq0.05~\mathrm{m},\nonumber\\
\left|v_{\mathrm{corr},k}-\hat v_{\mathrm{rel},k}^{\mathrm{MEKF}}\right|
&\leq0.25~\mathrm{m/s}.
\label{eq:lag_correction_guards}
\end{align}
Otherwise, the output reverts to the MEKF estimate. Accepted outputs are additionally rate limited to $1.0~\mathrm{m/s}$ in position and $3.5~\mathrm{m/s^2}$ in velocity. The resulting $z_{\mathrm{rel}}$ and $v_{\mathrm{rel}}$ are used to reconstruct the platform state as $\hat z_P^W=z_U^W-z_{\mathrm{rel}}$ and $\hat v_P^W=v_U^W-v_{\mathrm{rel}}$. This reconstructed state supplies the GRU predictor and the vision-based landing controller.

\section{GRU-Based Motion Prediction}
\label{prediction_module}

A gated recurrent unit (GRU)-based prediction module maps the recent history of reconstructed platform height to a future heave sequence. The reference frames, sign convention, and reconstruction $\hat z_P^W=z_U^W-z_{\mathrm{rel}}$ are defined once in Section~\ref{subsec:frames_notation}; therefore, a separate frame-definition subsection is not repeated here.

\subsection{Signal Conditioning}
The reconstructed platform height is normalized using
\begin{equation}
x(t)=\frac{\hat z_P^W(t)-z_b}{s},
\label{eq:normalized_heave_input}
\end{equation}
where $z_b$ is the nominal platform level and $s>0$ is the heave scale. For the reported experiments,
\begin{equation}
z_b=1.983~\mathrm{m},
\qquad
s=0.32~\mathrm{m}.
\label{eq:experimental_bias_scale_values}
\end{equation}
Equations~\eqref{eq:platform_height_reconstruction} and~\eqref{eq:normalized_heave_input} give the vision-based input
\begin{equation}
x(t)=
\frac{z_U^W(t)-z_{\mathrm{rel}}(t)-z_b}{s}.
\label{eq:vision_gru_input_general}
\end{equation}
It can be observed that $x(t)$ is dimensionless and approximately zero-centered. The same $z_b$ and $s$ used to normalize the training data are used for inference and output reconstruction. For deployment under a different nominal deck level or motion amplitude, these parameters may need to be recalibrated from an initial observation interval, provided that the resulting normalized signal remains within the training distribution.

\subsection{GRU Heave Predictor}
Let $x_k=x(k\Delta t)$ and let $N_h$ denote the length of input time history. At prediction time $k$, the oldest-to-newest input vector is
\begin{equation}
\mathbf{x}_k=
\begin{bmatrix}
x_{k-N_h+1} & \cdots & x_k
\end{bmatrix}^{T}.
\label{eq:gru_input_sequence}
\end{equation}
The trained GRU implements the sequence-to-sequence mapping
\begin{equation}
\hat{\mathbf{x}}_{k}^{+}
=f_{\theta}(\mathbf{x}_k)
=
\begin{bmatrix}
\hat x_{k+1\mid k} & \cdots & \hat x_{k+N_p\mid k}
\end{bmatrix}^{T},
\label{eq:gru_mapping}
\end{equation}
where $N_p$ is the prediction-horizon length and $\hat x_{k+i\mid k}$ denotes the normalized forecast for time $k+i$ issued at time $k$. The conditioning index $k$ distinguishes this forecast from one issued at a later sample $k'$.

The predictor operates at $f_s=20$~Hz, using a 40-s history and a 6-s prediction horizon. The 6-s horizon provides sufficient time for peak detection, landing-window selection, and descent planning while keeping forecast error and onboard computation within acceptable bounds. Hence,
\begin{equation}
N_h=40f_s=800,
\qquad
N_p=6f_s=120.
\label{eq:gru_horizon_samples}
\end{equation}
These horizons were selected empirically. Among the evaluated input lengths, a 40-s history gave acceptable accuracy while keeping the initial hover needed to fill the history buffer short. The prediction horizon had to be long enough to span peak detection, landing-window selection, and descent planning, but extending it beyond 6~s reduced accuracy over the first 3--4~s of the forecast, which is the portion most relevant to the terminal descent.

The normalized forecast is returned to the world-frame height as
\begin{equation}
\hat z_{P,k}^{W+}(i)
=s\hat x_{k+i\mid k}+z_b,
\qquad i=1,\ldots,N_p.
\label{eq:denormalized_prediction_absolute}
\end{equation}
This sequence is the only prediction output that is passed on to the landing-window estimator.

\subsection{Training Data and Objective}
The training data comprised four heave-motion records derived from the SCONE dataset~\cite{sconedataset}, corresponding to sea states 1--3 and each containing 36{,}000 samples at 20~Hz. These four records were used for parameter learning, while a fifth record was retained for validation. Each example paired an $N_h=800$ sample history with an $N_p=120$ sample target.

Let $\mathbf y=[y_1,\ldots,y_{N_p}]^T$ and $\hat{\mathbf y}=[\hat y_1,\ldots,\hat y_{N_p}]^T$ denote the target and predicted normalized sequences. To emphasize the first $T_w$ seconds, define
\begin{equation}
K_w=\min\!\left(\left\lfloor f_sT_w\right\rfloor,N_p-1\right),
\qquad 1\leq K_w<N_p,
\label{eq:early_weighted_samples}
\end{equation}
and
\begin{equation}
\mathcal L_{\mathrm{pos}}
=
\frac{w}{K_w}\sum_{j=1}^{K_w}(\hat y_j-y_j)^2
+
\frac{1}{N_p-K_w}\sum_{j=K_w+1}^{N_p}(\hat y_j-y_j)^2.
\label{eq:weighted_position_loss}
\end{equation}
This is the explicit sample-wise form of the two weighted mean-square-error terms.

Continuity is evaluated by appending the last $L$ observations $\mathbf o=[o_{-L+1},\ldots,o_0]^T$ to the target and the prediction. Let $\mathbf q=[\mathbf o^T,\mathbf y^T]^T$ and $\hat{\mathbf q}=[\mathbf o^T,\hat{\mathbf y}^T]^T$. Over the forecast portion, define $\Delta q_j=q_j-q_{j-1}$ and $\Delta^2q_j=q_j-2q_{j-1}+q_{j-2}$, where $q_0=o_0$ and $q_{-1}=o_{-1}$; define the predicted differences analogously. With
\begin{equation}
\alpha_j=
\begin{cases}
w, & 1\leq j\leq K_w,\\
1, & K_w<j\leq N_p,
\end{cases}
\end{equation}
the derivative-matching losses over the forecast portion are
\begin{equation}
\mathcal L_{\mathrm{vel}}
=\frac{1}{N_p}\sum_{j=1}^{N_p}
\alpha_j(\Delta\hat q_j-\Delta q_j)^2,
\end{equation}
\begin{equation}
\mathcal L_{\mathrm{acc}}
=\frac{1}{N_p}\sum_{j=1}^{N_p}
\alpha_j(\Delta^2\hat q_j-\Delta^2 q_j)^2.
\end{equation}
The separately weighted boundary-increment error is
\begin{equation}
\mathcal L_{\mathrm{jump}}
=\left[(\hat y_1-o_0)-(y_1-o_0)\right]^2
=(\hat y_1-y_1)^2.
\label{eq:jump_loss}
\end{equation}
The equality makes explicit that this term increases the weight on the first forecast sample; it is not an independent smoothness measure.

The complete objective is
\begin{equation}
\mathcal L=\mathcal L_{\mathrm{pos}}+
\beta_e\!\left(
\lambda_{\mathrm{jump}}\mathcal L_{\mathrm{jump}}+
\lambda_{\mathrm{vel}}\mathcal L_{\mathrm{vel}}+
\lambda_{\mathrm{acc}}\mathcal L_{\mathrm{acc}}
\right),
\end{equation}
where the continuity contribution is introduced over $E_{\mathrm{warm}}$ epochs using
\begin{equation}
\beta_e=\min\!\left(1,\frac{e}{E_{\mathrm{warm}}}\right).
\end{equation}
The retained configuration used $\lambda_{\mathrm{jump}}=8.0$, $\lambda_{\mathrm{vel}}=3.0$, $\lambda_{\mathrm{acc}}=0.8$, $L=5$, and $E_{\mathrm{warm}}=80$.

\subsection{Model Architecture Selection}
Several architectures were evaluated by varying the number of GRU layers and hidden units. Table~\ref{tab:gru_architecture_selection} reports the shortlisted models evaluated on the validation record. Prediction times were measured on a desktop NVIDIA RTX~5090 and are included only for relative comparison; they do not represent onboard latency. Errors in Table~\ref{tab:gru_architecture_selection} are computed in normalized units and reported in centimetres for a nominal $\pm25$~cm heave amplitude.

\begin{table*}[!t]
    \centering
    \small
    \caption{Representative GRU configurations evaluated during model-architecture selection.}
    \label{tab:gru_architecture_selection}
    \resizebox{\linewidth}{!}{%
    \begin{tabular}{lccccc}
        \hline
        \textbf{Hidden Sizes} & \textbf{Trainable Parameters} & \textbf{Avg. 3 s Error (cm)} & \textbf{Avg. 4 s Error (cm)} & \textbf{Full-Horizon Error (cm)} & \textbf{Desktop Time (ms)} \\
        \hline
        {[1024,1024]} & 9,575,544 & 2.952 & 3.130 & 3.742 & 0.305 \\
        {[2048,1024]} & 22,167,672 & 3.383 & 3.556 & 4.017 & 0.605 \\
        {[512,512,512,512]} & 5,580,408 & 3.669 & 3.886 & 4.389 & 0.708 \\
        {[512,512,512]} & 4,004,472 & 3.682 & 3.895 & 4.407 & 0.638 \\
        {[512,256,128]} & 1,546,104 & 3.665 & 3.960 & 4.540 & 0.661 \\
        {[1024,512,256]} & 6,139,512 & 4.176 & 4.354 & 4.734 & 0.663 \\
        {[512,512,512,128]} & 4,204,920 & 4.121 & 4.321 & 4.769 & 0.716 \\
        {[1024,512,256,128]} & 6,272,376 & 3.983 & 4.234 & 4.834 & 0.828 \\
        \hline
    \end{tabular}}
\end{table*}

The $[1024,1024]$ model was selected because it achieved the lowest reported errors while using fewer parameters than the larger $[2048,1024]$ alternative. Its average errors were 2.95~cm for the first 3~s, 3.13~cm for 4~s prediction, and 3.74~cm for the full horizon. The selected model was then deployed in the online prediction pipeline; its flight-test performance is evaluated in Section~\ref{experiments}.

\section{Landing Algorithm}
\label{landing_algorithm}

\begin{figure}[t]
    \centering
    \includegraphics[width=\linewidth]{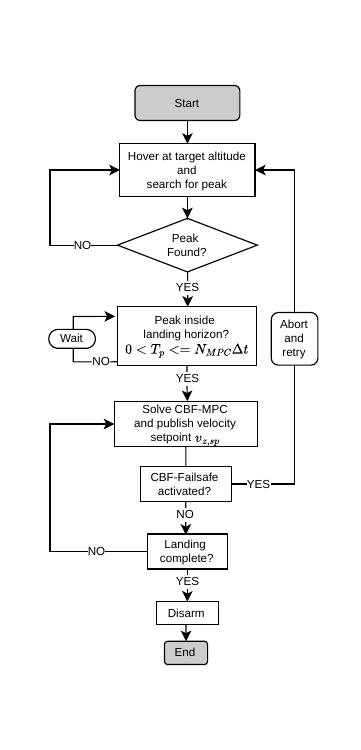}

    \caption{State-machine logic for peak selection, descent, rejection, and recovery.}
    \label{fig:heave_landing_state_machine}
\end{figure}

The landing algorithm combines predicted-peak selection, receding-horizon vertical trajectory generation, and a control barrier function (CBF). Touchdown is targeted near a local heave maximum, where the vertical velocity of the platform is small, and its elevated position reduces the required descent distance relative to a local minimum. The state machine in Fig.~\ref{fig:heave_landing_state_machine} supervises the peak search, descent, rejection, and recovery.

\subsection{Landing-Window Estimation}
At time $t_k$, the predictor supplies the world-frame platform-height forecast from~\eqref{eq:denormalized_prediction_absolute}, written compactly as
\begin{equation}
\hat{\mathbf z}_{P,k}^{W+}
=\begin{bmatrix}
\hat z_{P,k}^{W+}(1)&\cdots&\hat z_{P,k}^{W+}(N_p)
\end{bmatrix}^{T}.
\label{eq:landing_predicted_heave_vector}
\end{equation}
A forward moving average suppresses short-duration prediction fluctuations:
\begin{equation}
\bar z_{P,k}^{W+}(j)=\frac{1}{N_w}
\sum_{r=0}^{N_w-1}\hat z_{P,k}^{W+}(j+r),
\quad j=1,\ldots,N_p-N_w+1.
\label{eq:landing_moving_average}
\end{equation}
The implementation uses $N_w=7$, corresponding to 0.35~s at $f_s=20$~Hz, and applies SciPy's \texttt{find\_peaks} routine to the final $N_s=20$ smoothed samples. Minimum-prominence and minimum-separation checks reject numerical fluctuations and repeated detections of the same maximum. The earliest valid maximum is selected.

Let $j_p$ be the index of the earliest maximum in the smoothed sequence. Because~\eqref{eq:landing_moving_average} associates $\bar z(j)$ with a forward window, its center-aligned forecast index is
\begin{equation}
i_p=j_p+\left\lfloor\frac{N_w}{2}\right\rfloor.
\label{eq:landing_peak_index}
\end{equation}
The selected height and remaining time are
\begin{equation}
z_{\mathrm{peak}}=\bar z_{P,k}^{W+}(j_p),
\qquad
T_{\mathrm{peak}}=i_p\Delta t,
\label{eq:landing_peak_height_time}
\end{equation}
where $\Delta t=1/f_s=0.05$~s. These quantities define the terminal target and its timing for the trajectory generator.

\subsection{Vertical MPC Trajectory Generation}
Consistent with Section~\ref{subsec:frames_notation}, all vertical states use the upward-positive world-frame convention. The MPC state and input are
\begin{equation}
\mathbf x_i=\begin{bmatrix}z_{U,i}^{W}&v_{U,i}\end{bmatrix}^{T},
\qquad u_i=a_{U,i},
\label{eq:landing_mpc_state}
\end{equation}
and the forward-Euler model is
\begin{equation}
z_{U,i+1}^{W}=z_{U,i}^{W}+v_{U,i}\Delta t,
\qquad
v_{U,i+1}=v_{U,i}+u_i\Delta t.
\label{eq:landing_mpc_dynamics}
\end{equation}
Acceleration is the optimization input, but the low-level flight controller receives the optimized velocity setpoint.

Let $N$ denote the MPC horizon and define the number of steps to the selected peak without reusing the GRU horizon symbol $N_p$:
\begin{equation}
n_{\mathrm{peak}}
=\min\!\left(N,\max\!\left(1,
\left\lceil\frac{T_{\mathrm{peak}}}{\Delta t}\right\rceil\right)\right).
\label{eq:landing_steps_to_peak}
\end{equation}
The reference descends linearly to the selected peak and is then held constant:
\begin{equation}
\begin{aligned}
z_{\mathrm{ref},i}&=z_{U,0}^{W}+\eta_i
\left(z_{\mathrm{peak}}-z_{U,0}^{W}\right),\\
\eta_i&=\min\!\left(\frac{i}{n_{\mathrm{peak}}},1\right),
\quad i=0,\ldots,N.
\end{aligned}
\label{eq:landing_mpc_reference}
\end{equation}
The stage and terminal costs are defined as
\begin{equation}
\begin{aligned}
\ell_i={}&w_z(z_{U,i}^{W}-z_{\mathrm{ref},i})^2
+w_vv_{U,i}^2+w_uu_i^2\\
&+w_f(i)(z_{U,i}^{W}-z_{\mathrm{peak}})^2,\\
\ell_N={}&w_z(z_{U,N}^{W}-z_{\mathrm{ref},N})^2
+w_vv_{U,N}^2\\
&+w_f(N)(z_{U,N}^{W}-z_{\mathrm{peak}})^2.
\end{aligned}
\label{eq:landing_mpc_costs}
\end{equation}
At each control update, the nominal problem is
\begin{equation}
\begin{aligned}
\min_{\{u_i\}_{i=0}^{N-1}}\quad
&\sum_{i=0}^{N-1}\ell_i+\ell_N\\
\mathrm{s.t.}\quad
&\text{dynamics~\eqref{eq:landing_mpc_dynamics}},\\
&v_{\min}\leq v_{U,i}\leq v_{\max},\\
&u_{\min}\leq u_i\leq u_{\max},\\
&\mathbf x_0=
\begin{bmatrix}z_U^W(t_k)&v_U(t_k)\end{bmatrix}^{T}.
\end{aligned}
\label{eq:landing_mpc_problem}
\end{equation}
Here, $w_z$, $w_v$, and $w_u$ penalize altitude error, vertical speed, and control effort, respectively, while $w_f(i)$ emphasizes convergence to the selected platform height. The problem is resolved after every state and prediction update, and
\begin{equation}
v_{z,\mathrm{cmd}}=v_{U,1}^{\star}
\label{eq:landing_velocity_command}
\end{equation}
is sent to the flight controller.

\subsection{CBF Safety Constraint}
\begin{figure}[t]
    \centering
    \includegraphics[width=0.8\linewidth]{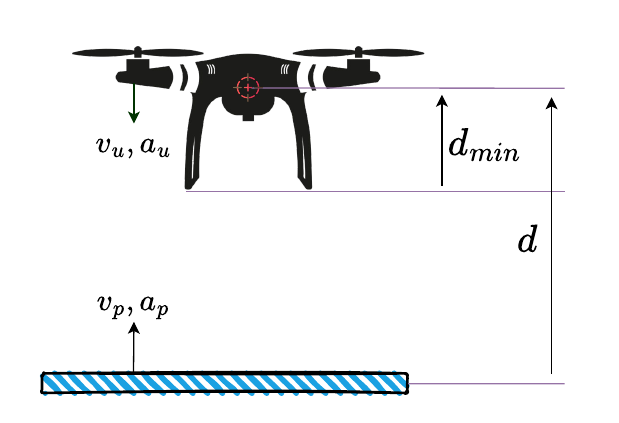}
    \caption{Vertical clearance and relative-motion quantities used by the CBF.}
    \label{fig:cbf_vertical_landing_safety}
\end{figure}

The CBF augments the proposed trajectory generator with a stopping-distance safety condition, following standard CBF theory~\cite{ames2017cbf,ames2019cbf}. Using the definitions from Section~\ref{subsec:frames_notation},
\begin{equation}
d=z_U^W-z_P^W,\qquad v_r=\dot d=v_U-v_P,
\label{eq:landing_relative_state}
\end{equation}
where $v_r<0$ denotes closing motion. The relative dynamics are
\begin{equation}
\dot d=v_r,\qquad \dot v_r=u-a_P.
\label{eq:landing_relative_dynamics}
\end{equation}
Let $d_{\min}>0$ be the height of the UAV center at landing-gear contact, $v_L\geq0$ the admissible touchdown magnitude for relative speed, and $a_{\mathrm{rel,max}}>0$ a conservative relative bound for braking. A differentiable approximation of the closing-speed magnitude is
\begin{equation}
\nu(v_r)=\frac{1}{2}\left(\sqrt{v_r^2+\varepsilon}-v_r\right),
\qquad \varepsilon>0,
\label{eq:landing_smooth_closing_speed}
\end{equation}
which approximates $\max(0,-v_r)$. Define the signed braking-distance term and barrier as
\begin{equation}
d_{\mathrm{stop}}=
\frac{\nu^2-v_L^2}{2a_{\mathrm{rel,max}}},
\qquad
h=d-d_{\min}-d_{\mathrm{stop}}.
\label{eq:landing_barrier}
\end{equation}
For $\nu\geq v_L$, $d_{\mathrm{stop}}$ is the distance required to reduce the closing-speed magnitude to $v_L$; when $\nu<v_L$, it is nonpositive and introduces no additional positive braking-distance requirement. Thus, $h\geq0$ requires the available clearance to satisfy the corresponding braking margin. Define
\begin{equation}
\nu'(v_r)=\frac{1}{2}\left(
\frac{v_r}{\sqrt{v_r^2+\varepsilon}}-1\right).
\end{equation}
The correct barrier derivative is then
\begin{equation}
\dot h=v_r-
\frac{\nu\nu'}{a_{\mathrm{rel,max}}}(u-a_P).
\label{eq:landing_barrier_derivative}
\end{equation}
The MPC enforces the standard CBF condition,
\begin{equation}
\dot h_0+\gamma h_0\geq0,
\label{eq:landing_cbf_constraint}
\end{equation}
at $i=0$ for every MPC tick using the current estimates of $z_P^W$, $v_P$, and $a_P$. Enforcing the constraint at the first stage of every receding-horizon update restricts commands that would produce insufficient braking clearance. The implemented CBF parameters are given in Table~\ref{tab:landing_controller_parameters}. Figure~\ref{fig:cbf_vertical_landing_safety} illustrates the geometry.

The MoCap and vision experiments used the same controller formulation but separate fixed parameter sets. The vision configuration limited the maximum descent speed, increased the vertical-velocity penalty, and extended the terminal velocity penalty to accommodate the lower measurement rate, latency, and uncertainty of the onboard estimates. Each parameter set remained unchanged across the ten reported trials in its sensing group.

\begin{table*}[t]
\centering
\caption{Landing-controller parameters used in the hardware experiments.}
\label{tab:landing_controller_parameters}
\setlength{\tabcolsep}{4pt}
\renewcommand{\arraystretch}{1.12}
\resizebox{\textwidth}{!}{%
\begin{tabular}{@{}llll@{}}
\hline
\textbf{Group} & \textbf{Parameter} & \textbf{MoCap} & \textbf{Vision} \\
\hline
MPC discretization & $\Delta t$; state points; shooting intervals & $0.05$~s; $90$; $89$ & $0.05$~s; $90$; $89$ \\
MPC bounds & $[v_{\min},v_{\max}]$; $[u_{\min},u_{\max}]$ & $[-0.6,1.5]$~m/s; $[-2,2]$~m/s$^2$ & $[-0.4,1.5]$~m/s; $[-2,2]$~m/s$^2$ \\
MPC cost & $w_z$; $w_v$; $w_u$; target-position weight & $10$; $1$; $0.1$; $50$ & $10$; $5$; $0.1$; $50$ \\
Terminal shaping & Additional velocity weight; tail points & $2400$; $40$ & $2400$; $50$ \\
CBF & $d_{\min}$; $a_{\mathrm{rel,max}}$; $\gamma$; $\varepsilon$; $v_L$ & $0.27$~m; $0.7$~m/s$^2$; $2.24$~s$^{-1}$; $10^{-4}$~(m/s)$^2$; $0.20$~m/s & $0.27$~m; $0.7$~m/s$^2$; $2.30$~s$^{-1}$; $10^{-4}$~(m/s)$^2$; $0.30$~m/s \\
State machine & Hover altitude; initial hover; recovery dwell & $3.2$~m; $10$~s; $4$~s & $3.2$~m; $10$~s; $4$~s \\
Landing transition & UAV-center-to-platform threshold & $0.36$~m & $0.36$~m \\
Solver & NLP; Hessian; QP solver; maximum iterations & SQP; exact; partial-condensing HPIPM; $100$ & SQP; exact; partial-condensing HPIPM; $100$ \\
\hline
\end{tabular}}
\end{table*}

In both configurations, the landing transition is activated when the estimated UAV-center-to-platform distance is below $0.36$~m. Since the nominal UAV-center-to-landing-gear offset is $d_{\min}=0.27$~m, this condition corresponds to 0.09~m, approximately 10~cm, of landing-gear clearance.

\subsection{Rejection and Recovery Logic}
A descent begins only when
\begin{equation}
0<T_{\mathrm{peak}}\leq N\Delta t.
\label{eq:landing_peak_horizon_condition}
\end{equation}
The UAV otherwise remains at hover until the peak enters the horizon; an elapsed peak is discarded. During descent, MPC infeasibility, violation of the CBF conditions, loss of a valid landing window, or expiration of that window triggers an abort. The UAV returns to its hover altitude, clears the rejected peak after a short dwell, and resumes the search for the next peak. Consequently, a landing command is accepted only when the target remains temporally reachable and both the optimization and safety constraint remain feasible.

\section{Experiments and Results}
\label{experiments}
Experiments evaluate the system at three levels: real-time heave prediction, peak selection, and closed-loop autonomous landing. MoCap measurements provide a low-noise reference, while onboard vision evaluates the practical perception-control pipeline. All reported vision-based prediction and landing experiments use the lag-compensated pipeline described in Section~\ref{subsec:lag_compensation}.

\subsection{Experimental Setup}
\label{subsec:experimental_setup}
\begin{figure}[t]
    \centering
    \includegraphics[width=\linewidth]{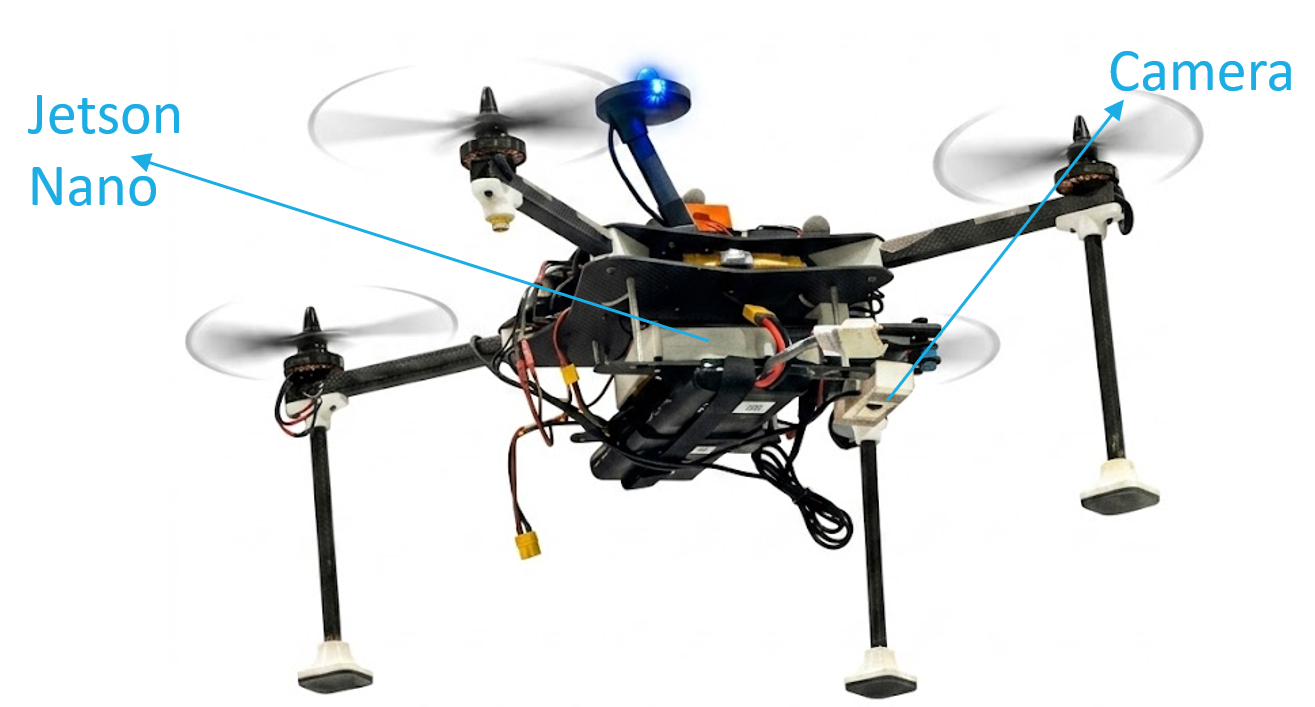}
    \caption{Quadrotor VTOL UAV and onboard sensing and computing hardware used in the heave-landing experiments.}
    \label{fig:heave_uav_setup}
\end{figure}

The aerial platform in Fig.~\ref{fig:heave_uav_setup} carries an NVIDIA Jetson Orin Nano developer kit (8-GB memory, 176-g mass), powered through a battery-elimination circuit. A downward-facing Logitech C922 camera supports 1080p acquisition at 30 frames/s and was operated at 720p and 24 frames/s; filtered pose and velocity estimates were published at 20~Hz. The onboard software performs marker detection, state estimation, GRU inference, landing-window selection, and trajectory generation, while the flight-control unit executes the commanded vertical-velocity setpoints using ArduPilot.

\begin{figure}[t]
    \centering
    \includegraphics[width=\linewidth]{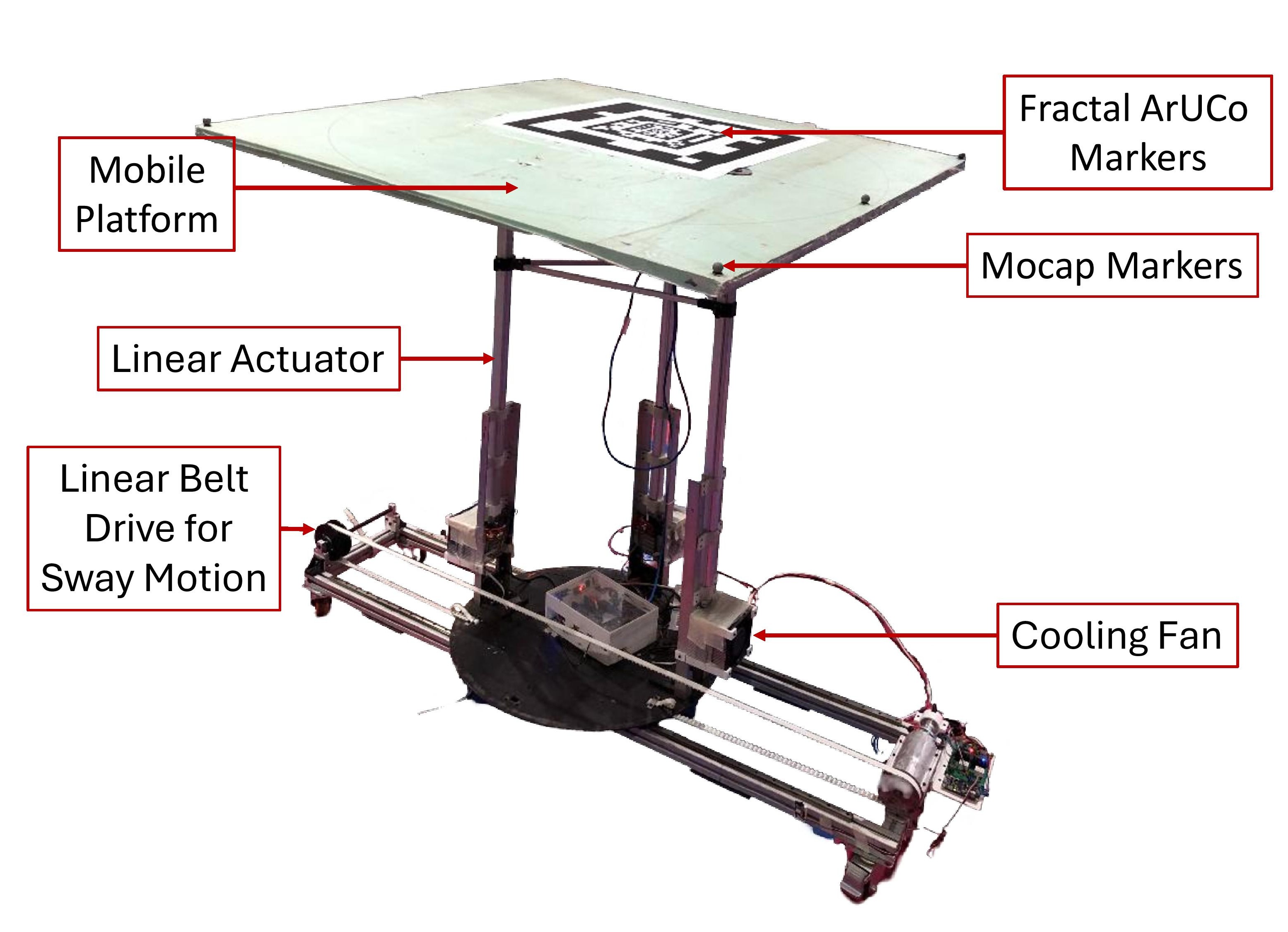}
    \caption{Four-DoF parallel-motion platform used in the experiments. Only heave is commanded in this work; the fractal marker provides vision-based deck-relative pose.}
    \label{fig:heave_platform_setup}
\end{figure}

The landing target shown in Fig.~\ref{fig:heave_platform_setup} is a four-degree-of-freedom 2PRU--1PRS parallel manipulator, where P, R, U, and S denote prismatic, revolute, universal, and spherical joints. It can reproduce heave, roll, pitch, and sway, and has a 75-cm actuator stroke with up to 70~cm of usable heave travel. The present work commands only heave; integration of the roll, pitch, and sway controllers with the proposed prediction-aided landing method is left for future work. The same commanded heave record was used for the MoCap- and vision-based experiments. This was the record retained for validation, as specified in Section~\ref{prediction_module}. Thus, the two sensing conditions were evaluated under the same source motion. The command spans $-0.32$ to $0.32$~m, its sample-wise vertical velocity ranges from approximately $-0.46$ to $0.43$~m/s, and its dominant period is approximately $8.9$~s. A fractal ArUco marker is mounted at the landing surface for vision-based relative pose estimation. The Qualisys motion-capture system records UAV and platform motion for ground-truth evaluation and provides the reference sensing condition used in the MoCap-based experiments.

\subsection{Validation of Heave Prediction}

The onboard GRU predictor was evaluated during hover above the moving platform. At 20~Hz, it generated $M=120$ samples spanning a 6~s horizon. Qualisys measurements provided the platform ground truth. MoCap-derived heave input established a reference with external pose feedback, whereas vision-derived input evaluated the onboard pipeline. Performance was assessed over complete forecast windows and at each prediction horizon.

\subsubsection{Window-Wise Forecast Accuracy}
\begin{figure*}[!t]
    \centering
    \subfloat[MoCap-based prediction.\label{fig:forecast_window_overlays}]{%
        \begin{minipage}{\textwidth}
        \centering
        \includegraphics[width=0.24\textwidth]{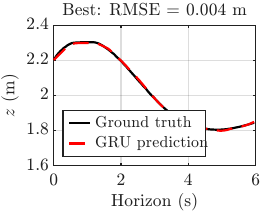}\hfill
        \includegraphics[width=0.24\textwidth]{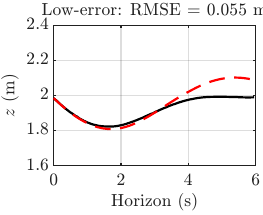}\hfill
        \includegraphics[width=0.24\textwidth]{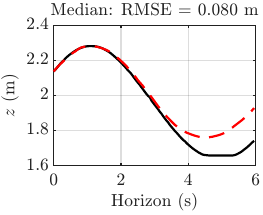}\hfill
        \includegraphics[width=0.24\textwidth]{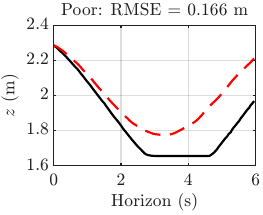}
        \end{minipage}}

    \subfloat[Vision-based prediction.\label{fig:forecast_window_overlays_vision_1}]{%
        \begin{minipage}{\textwidth}
        \centering
        \includegraphics[width=0.24\textwidth]{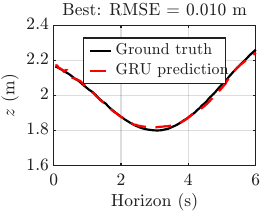}\hfill
        \includegraphics[width=0.24\textwidth]{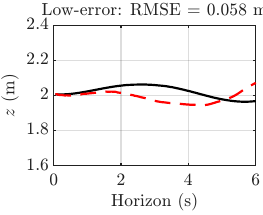}\hfill
        \includegraphics[width=0.24\textwidth]{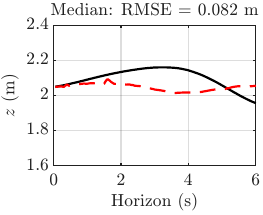}\hfill
        \includegraphics[width=0.24\textwidth]{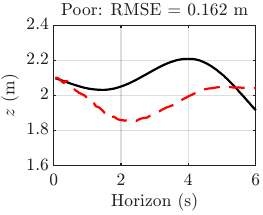}
        \end{minipage}}
    \caption{Representative 6~s heave prediction windows selected from the window-wise RMSE distribution. From left to right, the panels show the minimum-RMSE, 25th-percentile, median, and 95th-percentile cases.}
    \label{fig:forecast_window_comparison}
\end{figure*}

For the $k^{\mathrm{th}}$ forecast, the error at horizon index $i$ is
\begin{equation}
e_k(i)
=
\hat{z}_{P,k}^{W+}(i)
-
z_{P,\mathrm{gt}}^W(t_k+i\Delta t),
\qquad
i = 1,2,\ldots,M,
\label{eq:prediction_window_error}
\end{equation}
where \(\Delta t=0.05~\mathrm{s}\). Its window-wise RMSE is
\begin{equation}
\mathrm{RMSE}_k
=
\sqrt{
\frac{1}{M}
\sum_{i=1}^{M}
e_k(i)^2
}.
\label{eq:window_wise_rmse}
\end{equation}
Figure~\ref{fig:forecast_window_comparison} shows the minimum, 25th-percentile, median, and 95th-percentile RMSE windows, representing best, low-error, typical, and poor forecasts for each input. The predictions using MoCap as input track the heave more consistently, whereas the predictions using vision as input exhibit larger amplitude and phase errors. The distributional comparison follows below.

\begin{figure}[t]
    \centering
    \includegraphics[width=0.7\linewidth]{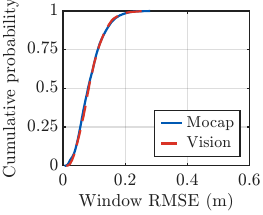}
    \caption{Empirical CDFs of window-wise RMSE over complete 6~s prediction windows for MoCap- and vision-based inputs.}
    \label{fig:cdf_prediction_rmse}
\end{figure}

The empirical cumulative distribution function (CDF) summarizes reliability across all $K$ valid windows:
\begin{equation}
F_{\mathrm{RMSE}}(r)
=
\frac{1}{K}
\sum_{k=1}^{K}
\mathbb{I}
\left(
\mathrm{RMSE}_k \leq r
\right).
\label{eq:rmse_cdf}
\end{equation}
As shown in Fig.~\ref{fig:cdf_prediction_rmse}, the MoCap and vision window-error distributions largely overlap. Table~\ref{tab:window_rmse_distribution} gives mean (median) RMSE values of 8.7 (8.0)~cm for MoCap input and 8.8 (8.2)~cm for vision input. Their 95th percentiles are 16.6 and 16.2~cm, respectively. Thus, the vision pipeline achieves similar aggregate accuracy over complete 6~s forecasts in these recorded hardware datasets.

\begin{table}[t]
\centering
\caption{Distribution of window-wise RMSE values for the 6~s heave prediction windows.}
\label{tab:window_rmse_distribution}
\setlength{\tabcolsep}{3pt}
\begin{tabular}{@{}lcc@{}}
\hline
\textbf{Metric} &
\shortstack{\textbf{MoCap-based}\\\textbf{prediction}} &
\shortstack{\textbf{Vision-based}\\\textbf{prediction}} \\
\hline
Number of windows          & 35806 & 12614 \\
Mean RMSE (cm)             & 8.7   & 8.8   \\
Median RMSE (cm)           & 8.0   & 8.2   \\
RMSE std. dev. (cm)        & 4.3   & 4.1   \\
Minimum RMSE (cm)          & 0.4   & 1.0   \\
25th percentile (cm)       & 5.5   & 5.8   \\
75th percentile (cm)       & 11.1  & 11.2  \\
95th percentile (cm)       & 16.6  & 16.2  \\
Maximum RMSE (cm)          & 28.0  & 27.0  \\
\hline
\end{tabular}
\end{table}

\subsubsection{Horizon-Wise Prediction Error}
To characterize error growth, the signed-error mean and standard deviation at horizon index $i$ are
\begin{equation}
\mu_e(i)
=
\frac{1}{K}
\sum_{k=1}^{K}
e_k(i),
\label{eq:horizon_mean_error}
\end{equation}
and
\begin{equation}
\sigma_e(i)
=
\sqrt{
\frac{1}{K-1}
\sum_{k=1}^{K}
\left(e_k(i)-\mu_e(i)\right)^2
}.
\label{eq:horizon_std_error}
\end{equation}
The corresponding magnitude measures are
\begin{equation}
\mathrm{RMSE}(i)
=
\sqrt{
\frac{1}{K}
\sum_{k=1}^{K}
e_k(i)^2
},
\label{eq:horizon_rmse}
\end{equation}
and
\begin{equation}
\mathrm{MAE}(i)
=
\frac{1}{K}
\sum_{k=1}^{K}
\left|e_k(i)\right|.
\label{eq:horizon_mae}
\end{equation}

\begin{figure}[t]
    \centering
    \subfloat[MoCap-based prediction.]{%
        \includegraphics[width=0.48\linewidth]{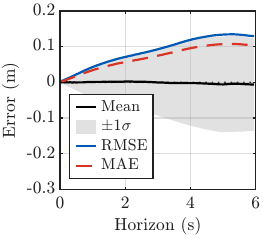}%
        \label{fig:horizon_error_mocap}}
    \hfill
    \subfloat[Vision-based prediction.]{%
        \includegraphics[width=0.48\linewidth]{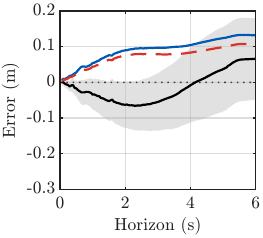}%
        \label{fig:horizon_error_vision_1}}
    \caption{Horizon-wise prediction errors for each sensing method. Each panel shows the mean error, \(\pm1\sigma\) spread, RMSE, and MAE over the 6~s forecast horizon.}
    \label{fig:horizon_error_mean_std}
    \label{fig:horizon_rmse_mae}
\end{figure}

Figure~\ref{fig:horizon_error_mean_std} shows increasing error with forecast horizon for both inputs. With MoCap, the mean signed error remains near zero, while RMSE rises to 9.3~cm at 3~s and 13.0~cm at 6~s. With vision, the mean signed error varies with horizon, crossing zero near 4~s; RMSE is 9.7~cm at 3~s and 13.3~cm at 6~s. Vision has higher RMSE at shorter horizons, but the curves cross between 3 and 4~s and remain close at 6~s. The vision result captures the complete onboard perception--prediction pipeline and supports evaluating both aggregate window error and horizon-dependent behavior when selecting future landing windows.

\subsection{Validation of Peak Estimator}

\begin{figure}[H]
    \centering
    \subfloat[Peak-height error.]{%
        \includegraphics[width=0.48\linewidth]{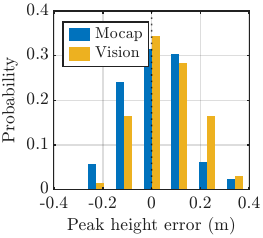}%
        \label{fig:peak_height_error_probability}}
    \hfill
    \subfloat[Platform velocity at the predicted peak time.]{%
        \includegraphics[width=0.48\linewidth]{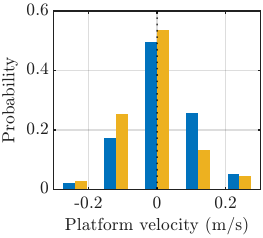}%
        \label{fig:platform_velocity_probability}}
    \caption{Probability-normalized peak-detection distributions for MoCap- and vision-based inputs. Bars for each common bin are shown side by side.}
    \label{fig:peak_detection_validation}
    \label{fig:mocap_peak_detection_validation}
    \label{fig:vision_peak_detection_validation}
\end{figure}

The landing-window estimator was evaluated against the ground-truth platform motion using the peak-height error
\begin{equation}
e_{z,\mathrm{peak}}
=
\hat{z}_{\mathrm{peak}}
-
z_{P,\mathrm{gt}}^W(\hat{t}_{\mathrm{peak}}),
\label{eq:peak_height_error}
\end{equation}
and the ground-truth platform velocity at the selected time,
\begin{equation}
v_{P,\mathrm{gt}}(\hat{t}_{\mathrm{peak}}).
\label{eq:platform_velocity_at_predicted_peak}
\end{equation}
An accurate selection should yield both a small height error and near-zero velocity, indicating proximity to the upper heave turning point. Because the MoCap and vision records contain different numbers of detections, each distribution in Fig.~\ref{fig:peak_detection_validation} is probability normalized.

For the 178 MoCap and 67 vision detections, the mean absolute peak-height errors are 10.0 and 10.9~cm, respectively. The corresponding mean absolute ground-truth platform velocities at the selected times are 0.075 and 0.071~m/s. These similar velocity magnitudes indicate that both inputs generally select low-velocity phases. The mean signed peak-height error is 1.6~cm for MoCap and 6.5~cm for vision, indicating a larger positive height bias in the vision detections. The distributions in Fig.~\ref{fig:peak_detection_validation} show the event-level variation behind these summary values.

The MoCap result provides a reference using external pose feedback, whereas the vision result characterizes onboard operation. Since both datasets contain peak-height and timing errors, each proposed window is screened by temporal rejection logic, the MPC feasibility test, and the CBF safety constraint before the vehicle commits to landing.

\subsection{Landing on Heaving Platform}

\begin{figure}[!t]
    \centering
    \includegraphics[width=\linewidth]{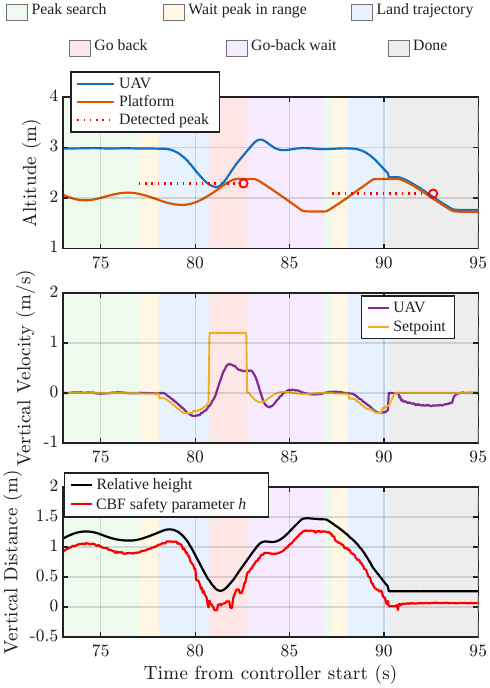}
    \caption{Representative vision-based heave-landing trial showing a rejected first approach followed by a completed second approach. The three panels show the UAV and platform altitudes; the measured UAV vertical velocity and commanded vertical-velocity setpoint; and the UAV-center-to-platform relative height together with the CBF safety parameter $h$. Background shading indicates the active controller state. During the first approach, $h<0$ triggers the abort and return to hover. During the second approach, $h$ remains nonnegative until the controller reaches the landing condition and issues the disarm command.}
    \label{fig:heave_landing_timeseries_plot}
\end{figure}

Closed-loop landing was evaluated in 20 trials: 10 with MoCap pose feedback and 10 with onboard vision. The MoCap trials provide a reference with external pose feedback, while the vision trials evaluate the complete perception--prediction--control pipeline. The groups used the same controller architecture and commanded heave record but the fixed sensing-specific parameter sets in Table~\ref{tab:landing_controller_parameters}. Consequently, this comparison evaluates the closed-loop performance achieved by each sensing configuration after tuning; it does not isolate the effect of replacing MoCap with vision under identical controller parameters.

Figure~\ref{fig:heave_landing_timeseries_plot} shows vision Trial~9. The first panel presents the UAV and platform altitudes, the second presents the measured UAV vertical velocity and its commanded setpoint, and the third presents the UAV-center-to-platform relative height and the CBF safety parameter $h$. The background colors indicate the active controller states. A predicted peak is marked by a circle, and the dashed line connects its detection time to its predicted occurrence time. Once a peak is detected, the controller transitions from \texttt{PEAK\_SEARCH} to \texttt{WAIT\_PEAK\_IN\_RANGE}, where it holds the UAV at the hover altitude until the predicted peak enters the MPC horizon. It then transitions to \texttt{LAND\_TRAJECTORY} and initiates the descent.

During the first approach, the predicted peak had approximately the correct phase but underestimated the actual peak height. Consequently, the available UAV-to-platform clearance became smaller than the required safety clearance plus stopping-distance margin, causing the CBF safety parameter $h$, defined in~\eqref{eq:landing_barrier}, to become negative. The resulting violation of the $h\geq0$ abort condition activated the failsafe, and the controller transitioned to \texttt{GO\_BACK}. The UAV returned to its prescribed hover altitude and entered \texttt{GO\_BACK\_WAIT}, where it was allowed to settle before the controller resumed \texttt{PEAK\_SEARCH}.

A second predicted peak was subsequently detected, and the controller repeated the \texttt{WAIT\_PEAK\_IN\_RANGE} and \texttt{LAND\_TRAJECTORY} transitions. Although both the phase and height of this prediction differed from the measured platform motion, the MPC reduced the commanded descent rate as the available braking margin decreased, and $h$ remained nonnegative throughout the second approach. When the landing-gear-to-platform clearance, computed as the UAV-center-to-platform relative height minus $d_{\min}$, decreased below the 10~cm landing threshold while the CBF condition remained satisfied, the controller transitioned to \texttt{DONE} and issued the disarm command.

\begin{figure*}[!t]
    \centering
    \subfloat[Vision estimation.\label{fig:landing_sequence_heaving_platform_a}]{%
        \includegraphics[width=0.23\textwidth,trim={250 1000 350 300},clip]{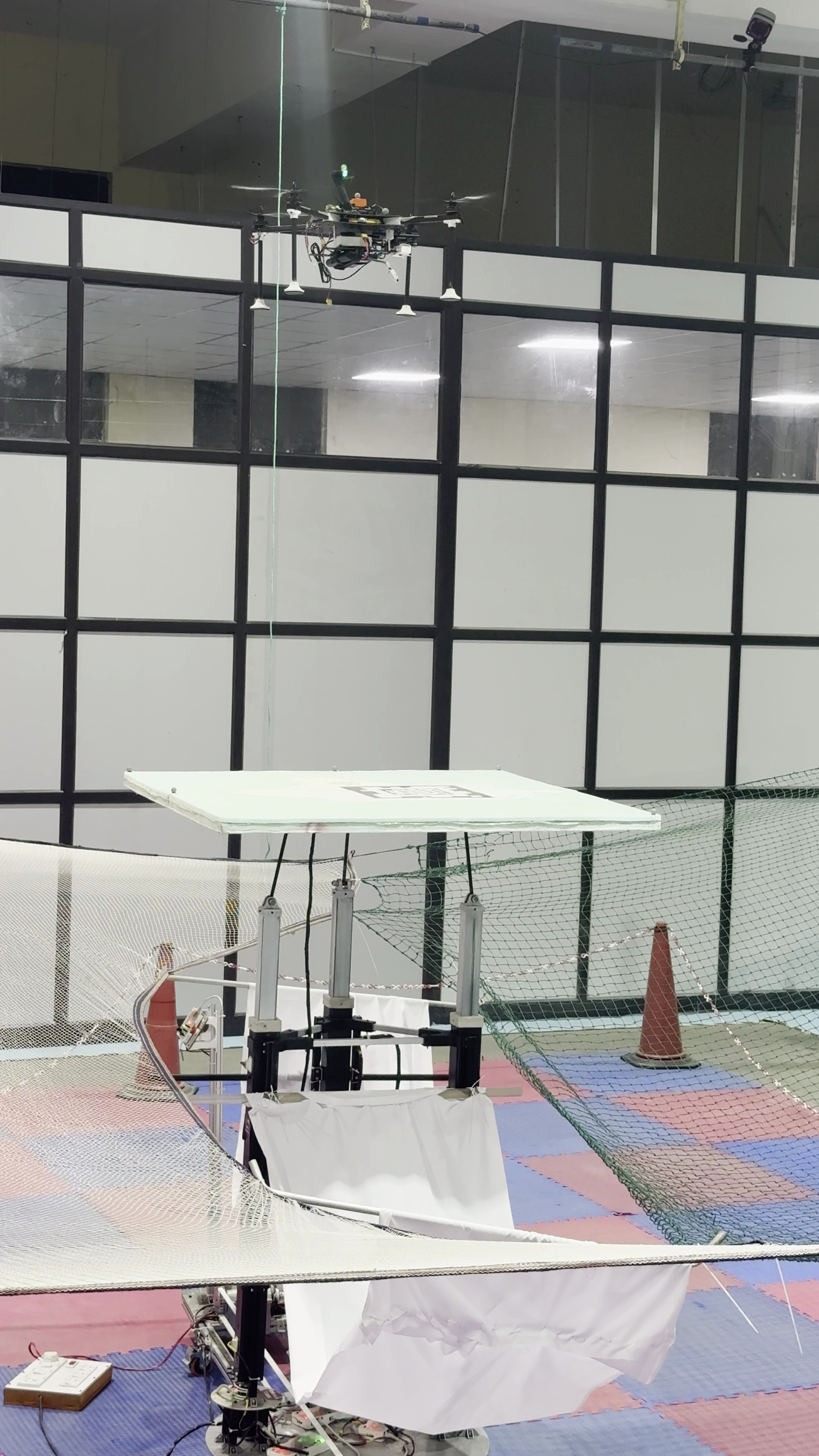}}
    \hspace{0.015\textwidth}
    \subfloat[GRU prediction.\label{fig:landing_sequence_heaving_platform_b}]{%
        \includegraphics[width=0.23\textwidth,trim={250 1000 350 300},clip]{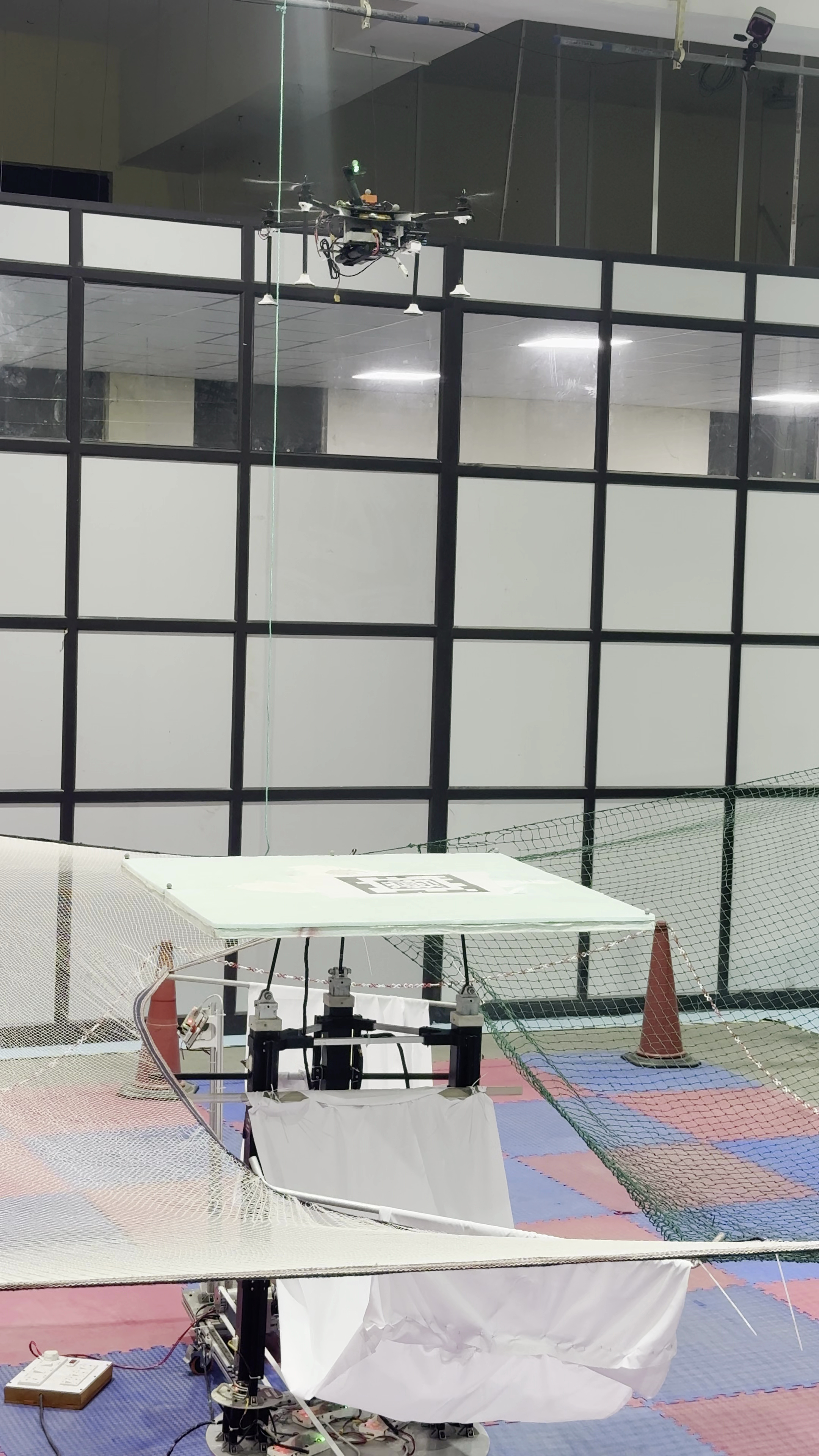}}
    \hspace{0.015\textwidth}
    \subfloat[$t=40$ s.\label{fig:landing_sequence_heaving_platform_c}]{%
        \includegraphics[width=0.23\textwidth,trim={250 1000 350 300},clip]{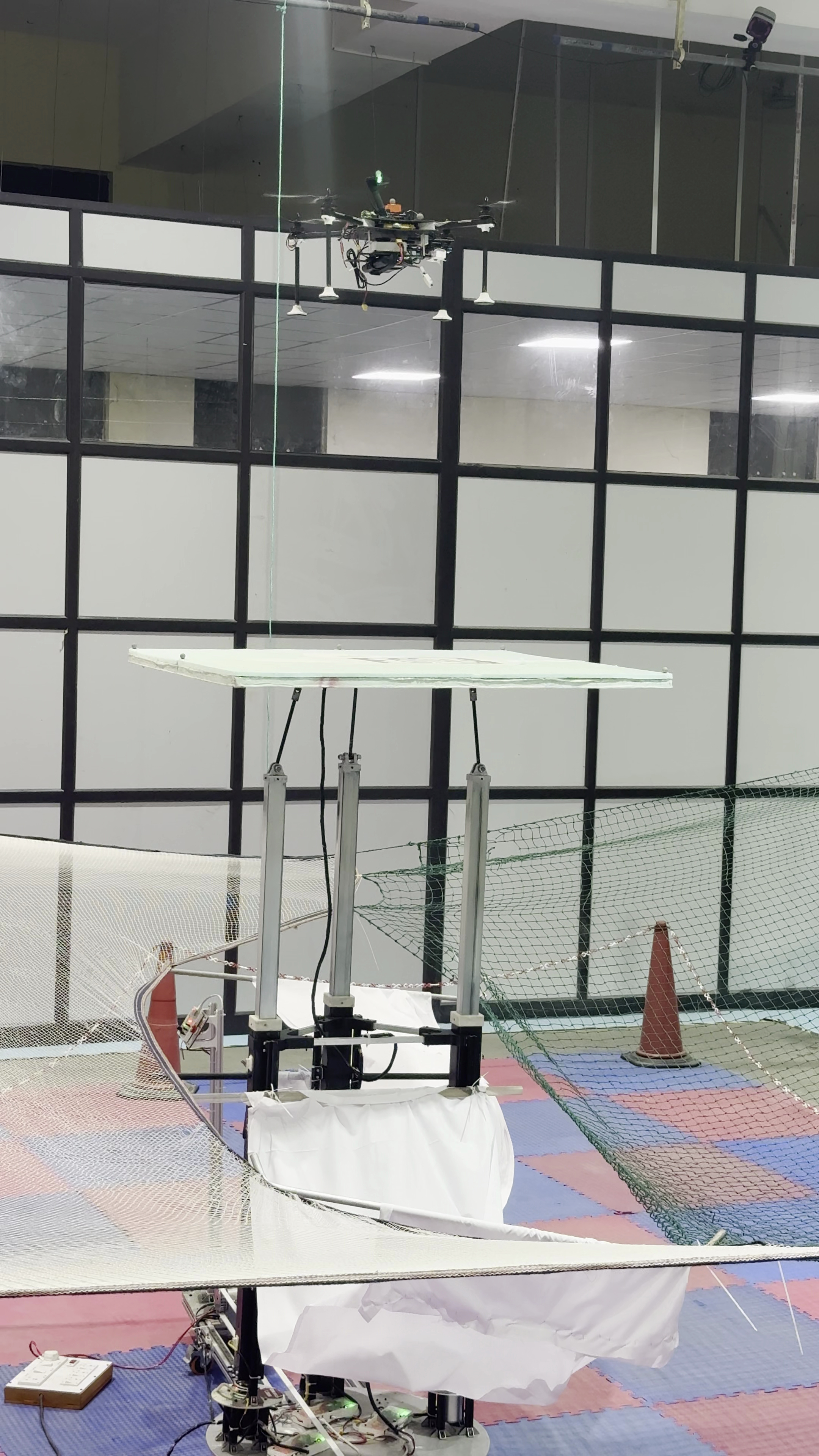}}

    \vspace{0.2em}

    \subfloat[MPC onset.\label{fig:landing_sequence_heaving_platform_d}]{%
        \includegraphics[width=0.23\textwidth,trim={250 1000 350 300},clip]{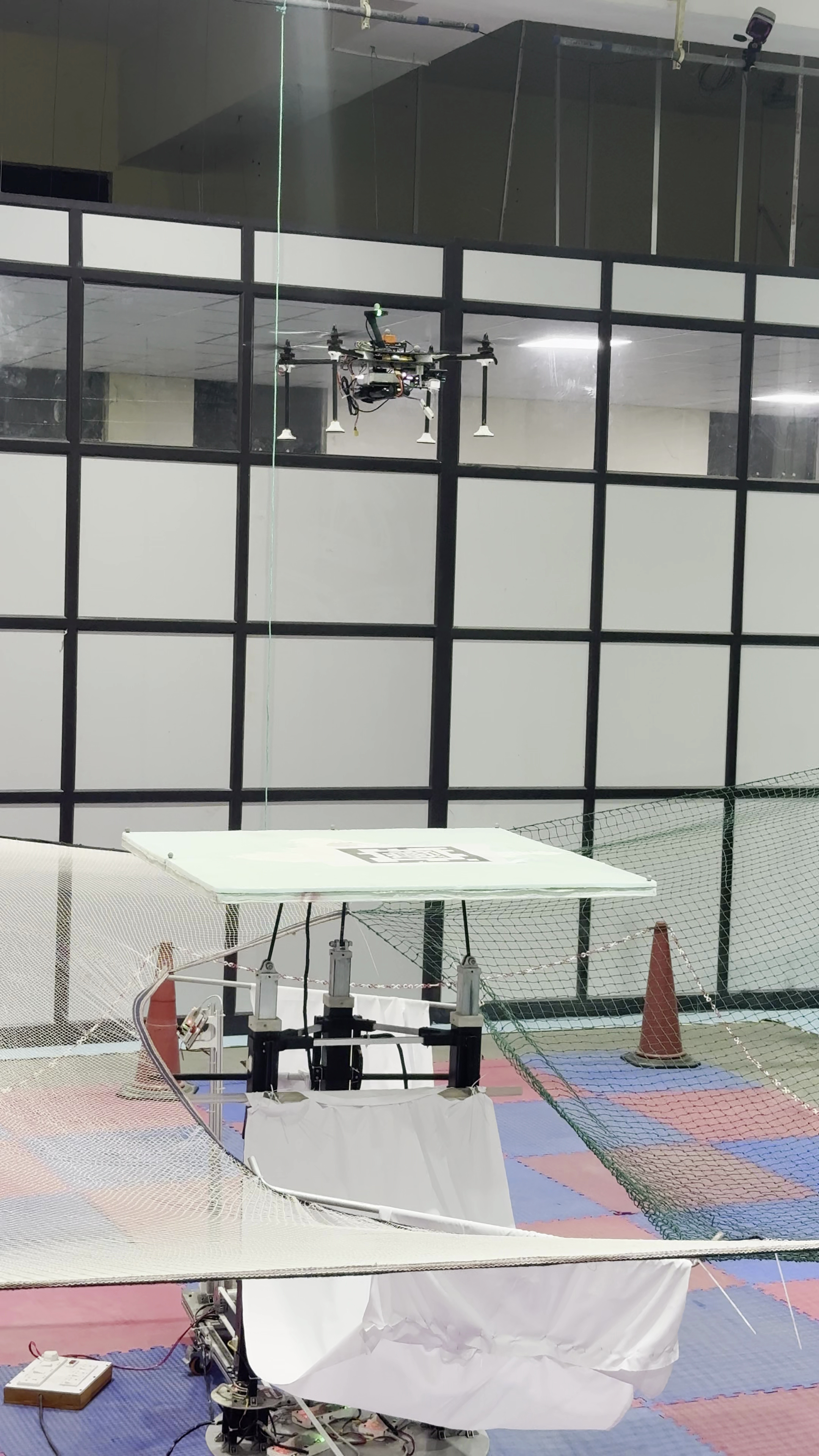}}
    \hspace{0.015\textwidth}
    \subfloat[Descent near peak.\label{fig:landing_sequence_heaving_platform_e}]{%
        \includegraphics[width=0.23\textwidth,trim={250 1000 350 300},clip]{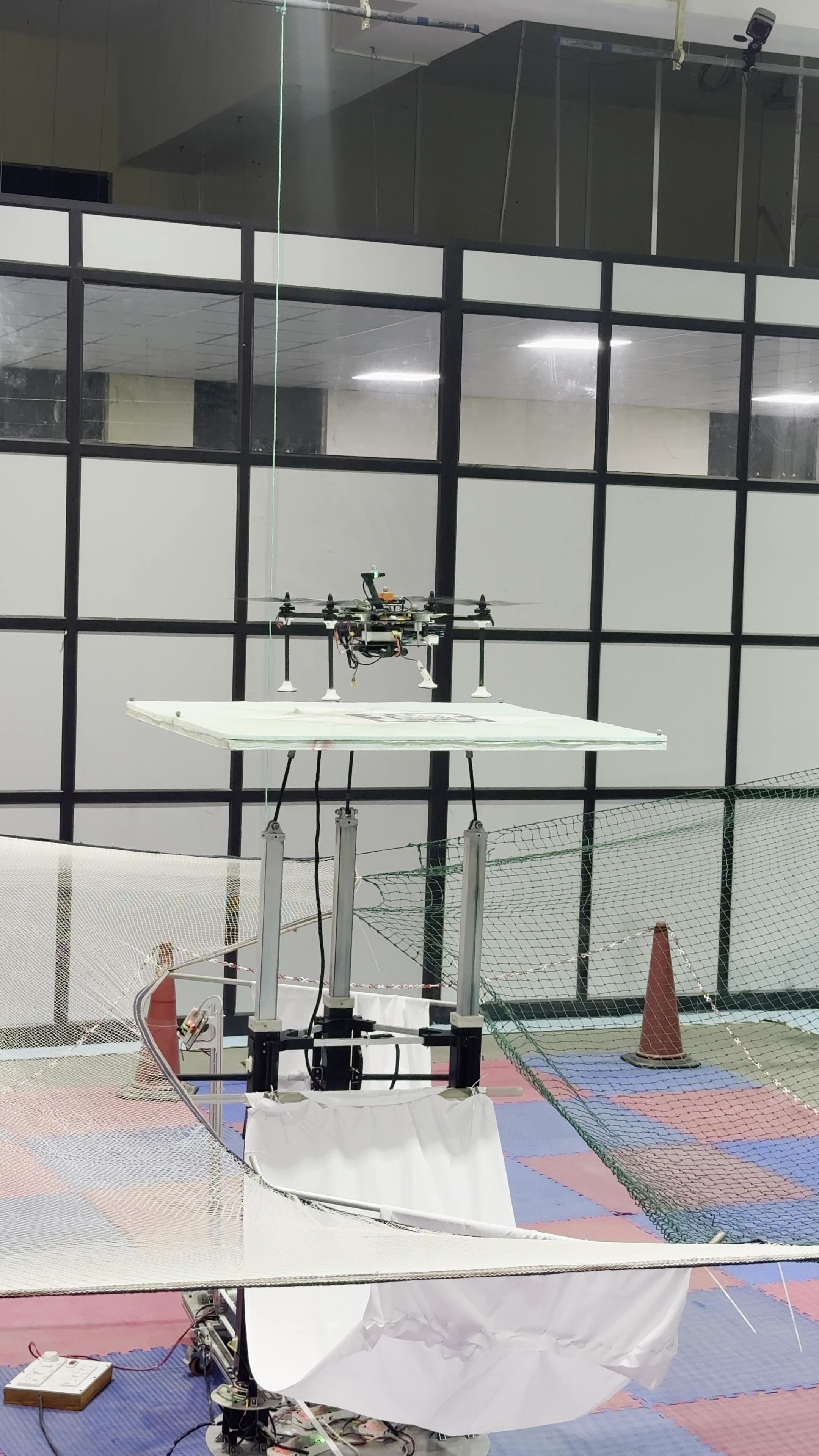}}
    \hspace{0.015\textwidth}
    \subfloat[Touchdown.\label{fig:landing_sequence_heaving_platform_f}]{%
        \includegraphics[width=0.23\textwidth,trim={250 1000 350 300},clip]{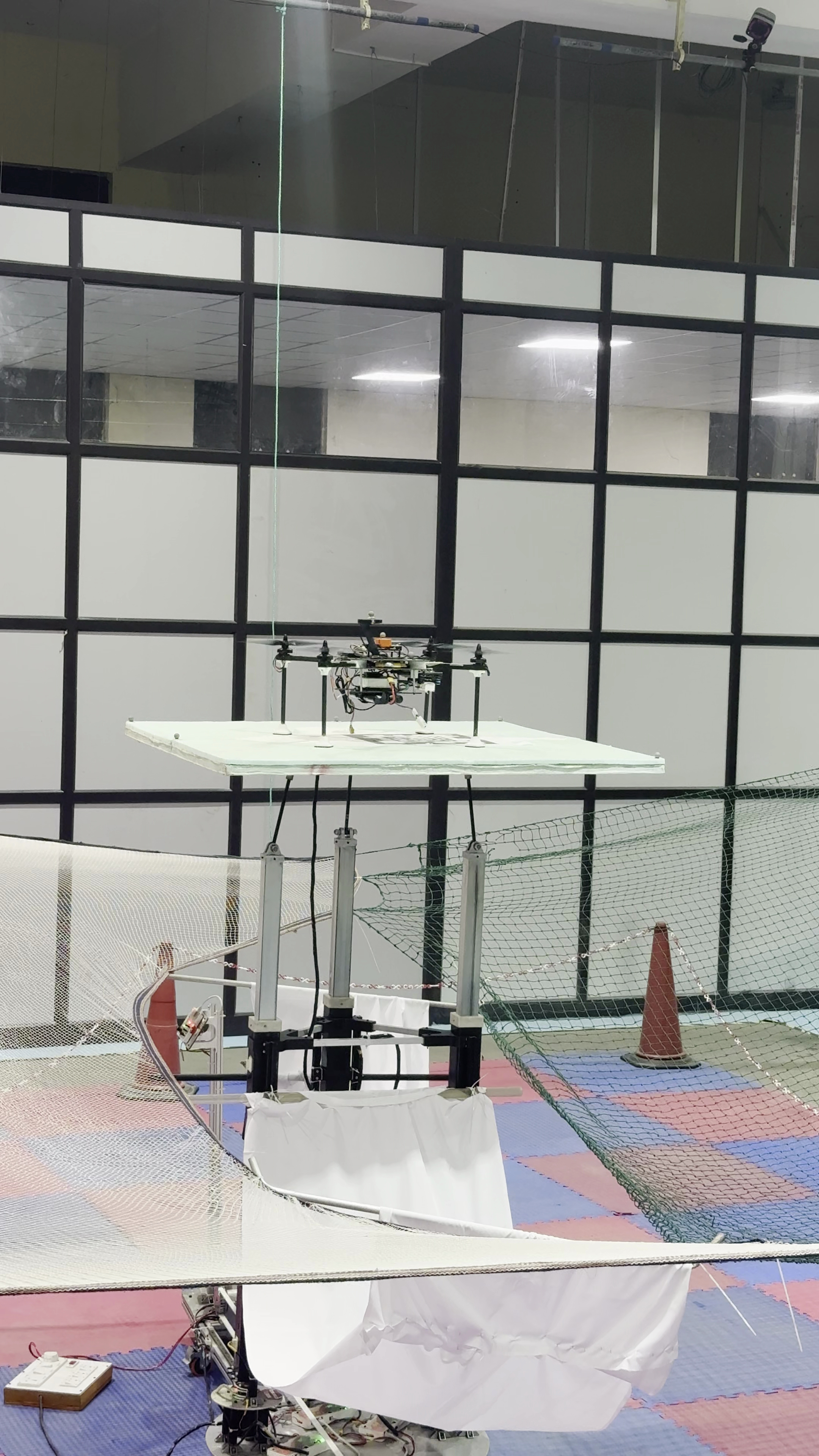}}
    \caption{Images of the UAV landing sequence on a platform simulating ship-deck heave motion.}
    \label{fig:landing_sequence_heaving_platform}
\end{figure*}

The 10~cm threshold was selected empirically during preliminary experiments to balance two observed effects. A smaller threshold increased the frequency of rejected approaches as the UAV descended closer to the moving platform, whereas a larger threshold caused disarm to occur farther above the platform and increased the subsequent relative speed at physical contact. The \texttt{DONE} transition therefore marks the controller termination and disarm event; the separately determined physical first-contact time is used for the touchdown-velocity analysis below. This trial illustrates a rejected first approach followed by a completed second approach through the controller's abort, recovery, peak-search, and reattempt sequence. Figure~\ref{fig:landing_sequence_heaving_platform} shows the physical stages from vision estimation and prediction through MPC onset, descent, and touchdown.

\setlength{\textfloatsep}{8pt}

\begin{figure}[hb]
    \centering
    \subfloat[Absolute relative vertical speed at disarm and first contact.]{\includegraphics[width=0.48\linewidth]{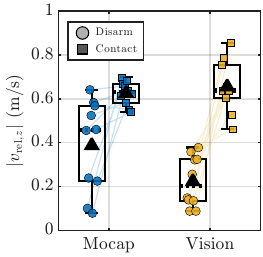}%
        \label{fig:td_rel_vel_boxplot}}
    \hfill
    \subfloat[Absolute platform vertical velocity at touchdown.]{\includegraphics[width=0.48\linewidth]{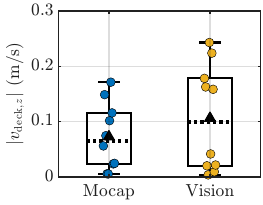}%
        \label{fig:td_platform_vel_boxplot}}
    \caption{Absolute relative vertical speed at disarm and first contact, and absolute platform vertical speed at first contact, for MoCap and vision trials. The dash-dot line denotes the median, and the black triangle denotes the mean.}
    \label{fig:touchdown_velocity_metrics}
\end{figure}

\begin{figure}[h]
    \centering
    \includegraphics[width=0.5\linewidth]{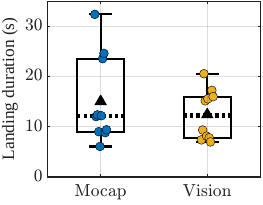}
    \caption{Landing-duration distributions for MoCap and vision trials. Duration is measured from the first effective \texttt{PEAK\_SEARCH} sample to the first contact. It includes peak search, any rejected approaches, and final descent. The dash-dot line denotes the median, and the black triangle denotes the mean.}
    \label{fig:landing_duration_boxplot}
\end{figure}

We evaluate absolute UAV-to-deck vertical speed at the controller
\texttt{DONE} state, when the disarm command is sent, and at physical
first contact. At either event time $t$,
\begin{equation}
    v_{\mathrm{rel},z}(t)
    =
    v_{\mathrm{UAV},z}(t)-v_{\mathrm{deck},z}(t).
\end{equation}
We denote the first-contact time by $t_{\mathrm{TD}}$. All touchdown quantities were computed from synchronized Qualisys ground-truth measurements. During the post-disarm descent, first contact was identified as the instant at which the landing-gear-to-platform clearance was within 0.5~cm of its minimum and the UAV reached its most negative vertical velocity immediately before its velocity converged to the platform velocity. The detected event was manually verified from the synchronized altitude and velocity histories of every trial. For reporting, a contact is classified as low speed when $|v_{\mathrm{rel},z}(t_{\mathrm{TD}})|<1~\mathrm{m/s}$. This reporting threshold is distinct from the controller parameter $v_L$ used in the CBF braking-distance calculation. The MoCap and vision controllers used $v_L=0.20~\mathrm{m/s}$ and $v_L=0.30~\mathrm{m/s}$, respectively, whereas the $1~\mathrm{m/s}$ criterion is applied only to classify the measured relative speed at physical first contact. This study-specific criterion is informed by the landing conditions reported for conventional multirotor landing gear in \cite{Tunney2026-wp}. The \texttt{DONE}/disarm speed is reported separately and does not determine this classification. The magnitude $|v_{\mathrm{deck},z}(t_{\mathrm{TD}})|$ indicates whether contact occurred near a low-velocity heave phase.

Figure~\ref{fig:touchdown_velocity_metrics} separates cutoff from physical contact. At disarm, the mean absolute relative speed is 0.388~m/s for MoCap and 0.224~m/s for vision (medians 0.458 and 0.202~m/s). At first contact, the means are 0.626 and 0.656~m/s (medians 0.631 and 0.638~m/s), respectively. Thus, the low speed at the vision cutoff does not imply equally low impact speed: the vehicle continues moving after disarm. The mean absolute deck speeds at contact are 0.073 and 0.106~m/s, respectively. The MoCap group had a slightly lower mean first-contact relative speed than the vision group. All 20 measured first-contact speeds were below the $1~\mathrm{m/s}$ low-speed touchdown criterion.

All ten vision trials therefore achieved low-speed touchdown with the vision-specific configuration, showing that the prediction, rejection, and CBF-constrained MPC architecture can be tuned for the measurement rate, latency, and uncertainty of the onboard sensing pipeline.

Figure~\ref{fig:landing_duration_boxplot} shows mean landing durations of 15.1~s for MoCap and 12.4~s for vision, with medians of 12.1 and 12.3~s. The nearly identical medians indicate comparable typical completion times under the tested conditions. The difference between the group means is influenced by the small sample sizes and the longer-duration trials and is not attributed to sensing alone.

\begin{table}[t]
\centering
\caption{Distribution of successful landings by attempt number for motion-capture and vision-based experiments.}
\label{tab:landing_attempt_distribution}
\renewcommand{\arraystretch}{1.2}
\setlength{\tabcolsep}{3pt}
\begin{tabular}{@{}lccc@{}}
\hline
\textbf{Method} & \textbf{Total trials} & \textbf{Attempt 1} & \textbf{Attempt 2} \\
\hline
MoCap  & 10  & 8 & 2  \\
Vision & 10  & 6 & 4  \\
\hline
\end{tabular}
\end{table}

Table~\ref{tab:landing_attempt_distribution} shows that MoCap completed eight of ten landings on the first attempt and two on the second; vision completed six on the first and four on the second. An attempt is one contiguous \texttt{LAND\_TRAJECTORY} interval. Every reported trial reached \texttt{DONE}, and none required a third attempt. Figure~\ref{fig:heave_landing_timeseries_plot} illustrates one of the reattempted vision landings.

\section{Conclusions}
\label{conclusions}

This work demonstrated the hardware feasibility of a heave-synchronized autonomous landing architecture combining GRU-based platform-motion prediction, future-peak selection, and CBF-constrained MPC descent control. The method schedules descent toward a predicted low-velocity heave phase and rejects a selected window when temporal-feasibility, clearance, or braking-margin requirements are violated. The experiments characterized MoCap- and vision-based configurations but did not compare the architecture against alternative predictors or landing controllers.

Over complete 6~s forecasts, MoCap and vision inputs produced similar mean window RMSE values of 8.7 and 8.8~cm (medians 8.0 and 8.2~cm), although the horizon-wise errors followed different trajectories. At the selected peak times, mean absolute ground-truth deck speeds were 0.075~m/s for MoCap and 0.071~m/s for vision. Mean absolute peak-height errors were 10.0 and 10.9~cm, while the vision detections had a larger positive signed height bias. These residual prediction and peak-selection errors motivate downstream feasibility and safety checks.

All 20 closed-loop trials achieved touchdown. Eight of ten MoCap trials and six of ten vision trials were completed on the first attempt; the others were completed on the second attempt. The mean absolute relative speed at first contact was 0.626~m/s for MoCap and 0.656~m/s for vision. All 20 contacts met the \(1~\mathrm{m/s}\) low-speed touchdown criterion. Mean durations from effective peak search to touchdown were 15.1 and 12.4~s, respectively, while the corresponding medians were 12.1 and 12.3~s. The MoCap- and vision-based trials therefore exhibited comparable median landing durations under the tested conditions. The vision representative trial illustrates a rejected approach followed by a successful reattempt. These observations distinguish successful controller termination from the physical touchdown conditions measured afterward. The sensing groups used separate fixed parameter sets; in particular, the vision configuration used a lower descent-speed bound and stronger velocity regulation to accommodate its lower measurement rate, latency, and uncertainty. Its successful trials demonstrate that the proposed architecture can be tuned for onboard sensing while retaining prediction-aided peak selection, safety-based rejection, and low-speed touchdown. The results establish feasibility for the complete onboard configuration under the tested conditions. It does not claim superiority over alternative predictors or landing controllers.

The conclusions are limited to controlled laboratory trials with vertical heave, a visible fiducial marker, and 20 landings. Future work should evaluate coupled heave, roll, pitch, and sway; more aggressive and less repeatable deck motions; partial or intermittent marker loss; and larger trial sets. Improving vision-state estimation and incorporating prediction uncertainty directly into landing-window selection and trajectory optimization are also important steps toward operation under realistic maritime conditions.

\bibliographystyle{IEEEtran}
\bibliography{references}

\end{document}